\documentclass[acmtog]{acmart}
\acmSubmissionID{1726}

\usepackage{booktabs} 
\usepackage{multirow}
\usepackage{cleveref}
\usepackage{amsmath}
\usepackage{graphicx}
\usepackage{wrapfig}
\usepackage{tabularx}
\usepackage{array}
\usepackage{color}
\usepackage{colortbl}

\usepackage{bbm}
\usepackage{overpic}

\usepackage[ruled]{algorithm2e} 

\SetAlFnt{\small}
\SetAlCapFnt{\small}
\SetAlCapNameFnt{\small}
\SetAlCapHSkip{0pt}

\acmJournal{TOG}

\newcommand{\rzz}[1]{\textcolor{black}{#1}}   
\newcommand{\old}[1]{\textcolor{black}{#1}}   
\newcommand{\rz}[1]{#1} 
\newcommand{\am}[1]{\textcolor{black}{#1}}

\newcommand{\saura}[1]{\textcolor{black}{#1}}

\definecolor{revcolor}{RGB}{0,0,0}
\newcommand{\rev}[1]{\textcolor{revcolor}{#1}}

\makeatletter
\newenvironment{revblock}
  {\color{revcolor}%
   \let\default@color\current@color            
   \everymath\expandafter{\the\everymath\color{revcolor}}%
   \everydisplay\expandafter{\the\everydisplay\color{revcolor}}}
  {}
\makeatother

\newcommand{\eg}{e.g.,}

\makeatletter
\@ifundefined{@titlefont}
  {\newcommand{\SMtitlefont}{\LARGE\bfseries}}
  {\newcommand{\SMtitlefont}{\@titlefont}}
\@ifundefined{@subtitlefont}
  {\newcommand{\SMsubtitlefont}{\large\itshape}}
  {\newcommand{\SMsubtitlefont}{\@subtitlefont}}
\makeatother

\copyrightyear{2026}
\acmYear{2026}
\setcopyright{cc}
\setcctype{by-nc-nd}
\acmConference[SA Conference Papers '26]{SIGGRAPH Asia 2026 Conference Papers}{December 01--04, 2026}{Kuala Lumpur, Malaysia}
\acmBooktitle{SIGGRAPH Asia 2026 Conference Papers (SA Conference Papers '26), December 01--04, 2026, Kuala Lumpur, Malaysia}
\acmDOI{10.1145/3829340.3842260}
\acmISBN{979-8-4007-2842-6/2026/12}

\begin{document}
\title{Functionalization via Structure Completion and Motion Rectification}


\author{Mingrui Zhao}
\affiliation{%
 \institution{Simon Fraser University}
 \country{Canada}}
\email{mza143@sfu.ca}
\author{Sai Raj Kishore Perla}
\affiliation{%
 \institution{Simon Fraser University}
 \country{Canada}
 }
 \author{Kai Wang}
 \affiliation{%
 \institution{Simon Fraser University}
 \country{Canada}
 }
 \affiliation{%
 \institution{ShanghaiTech University}
 \country{China}
 }
 \author{Sauradip Nag}
 \affiliation{%
 \institution{Simon Fraser University and Augmenta}
 \country{Canada}
 }
 \author{Duc Anh Nguyen}
 \affiliation{%
 \institution{Simon Fraser University}
 \country{Canada}
 }
 \author{Jiayi Peng}
 \affiliation{%
 \institution{Simon Fraser University}
 \country{Canada}
 }
 \author{Ruiqi Wang}
 \affiliation{%
 \institution{Simon Fraser University}
 \country{Canada}
 }
 \author{Angel X. Chang}
 \affiliation{%
 \institution{Simon Fraser University and Alberta Machine Intelligence Institute (Amii)}
 \country{Canada}
 }
 \author{Manolis Savva}
 \affiliation{%
 \institution{Simon Fraser University}
 \country{Canada}
 }
 \author{Ali Mahdavi-Amiri}
 \affiliation{%
 \institution{Simon Fraser University}
 \country{Canada}
 }
 \author{Hao Zhang}
 \affiliation{%
 \institution{Simon Fraser University and Augmenta}
 \country{Canada}
 }

\renewcommand\shortauthors{Zhao, M. et al}

\begin{abstract}

Acquisition and creation of 3D assets have been largely view- or appearance-driven. As a result, existing digital 3D models often lack the requisite structural components to function as intended, such as joints, supports, interiors, or interaction elements. At the same time, even human-annotated motions are frequently error-prone, leading to physically implausible behavior. We introduce object \emph{functionalization}, a novel task aimed at transforming visually plausible but non-functional 3D models into functional and physically operable ones.
\rzz{We formulate functionalization as a \emph{graph completion} problem over a new functional graph representation, where labeled nodes represent object parts, labeled edges encode functional and contact relations, and movable nodes carry motion attributes, so that structural functional deficiencies manifest as missing nodes or incorrect edges. We develop a \emph{neural Graph Functionalizer} (GraFu) to complete an incomplete graph representing a non-functional 3D object. The completed graph then drives a geometry realization stage that instantiates predicted connectors and structural elements in 3D, with the compelling side effect of rectifying erroneous human-annotated and predicted motions. To support training and evaluation, focusing on furniture as a rich and challenging target category, we introduce FurFun-233, a dataset of 233 paired non-functional and functionalized furniture models. On PartNet-Mobility (``zero-shot'') and HSSD test sets, our method matches state-of-the-art methods in motion prediction accuracy while substantially improving functionality in terms of collision and connectivity. \rev{Project page: \url{https://mingrui-zhao.github.io/Functionalization/}.}} 


\end{abstract}

\keywords{Shape editing, Functional modelling, Shape graphs, Object functionalization, Articulated 3D objects}

\begin{teaserfigure}
  \includegraphics[width=\textwidth]{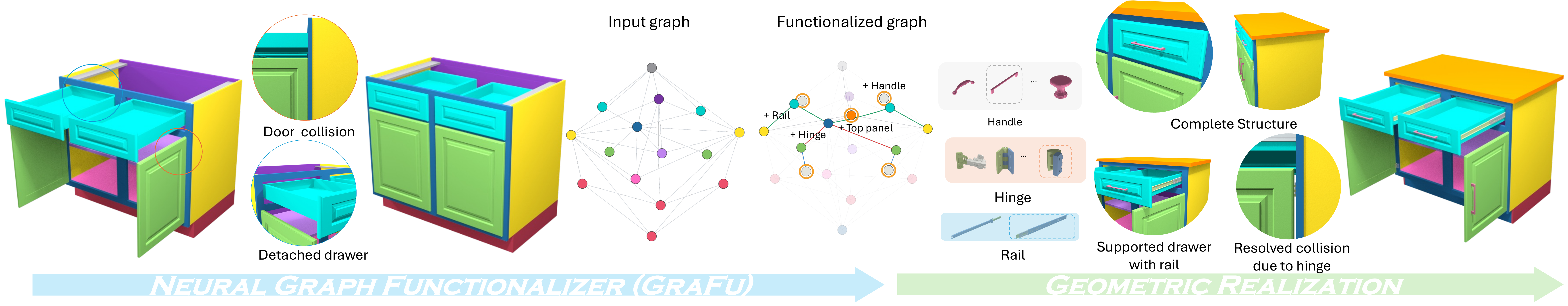}
  \caption{
  Our method \emph{functionalizes} a cabinet model that cannot function as it should, due to a) missing connectors for its door (hinge) or drawer (rails) to open and be held in place, and b) a collision between the door and the side panel. Our solution pipeline consists of a neural graph functionalizer (GraFu) for functional graph completion, e.g., adding missing connectors (new graph edges), as well as drawer handles (new graph nodes), followed by geometric realization to put actual hinges and rails in place. The hinge mechanism enforces a motion axis rectification that effectively fixes the initial collision.
  }
  \Description{Teaser}
  \label{fig:teaser}
\end{teaserfigure}

\maketitle

\section{Introduction}
\label{sec: Introduction}

We live in a 3D world, surrounded by 3D objects that we constantly act on and interact with to carry out our daily activities. These activities define how we function as humans, while the same also applies to the robots we design to assist us. Correspondingly, the 3D objects have been constructed with their intended functionalities to afford the actions of humans or robotic agents alike. For example, all cabinets must have interior shelving and tops for storage and support, drawers have handles or grooves so that they can be pulled, doors have hinges to swing open, while a swivel chair must have a full cylinder assembly to stabilize and rotate its seat.

In reality, however, very few, if any, of the existing 3D assets fully possess the necessary \emph{functional} components such as joints, handles, tracks, or other constituents of a part assembly for them to work. The majority of daily objects such as furniture and appliances do not have interior structures, and worse yet, a great many of them contain self collisions, under- or over-segmented parts, misalignments, etc., which prevent realization of functional motions. Even when the assets were manually created by 3D artists~\cite{collins2022abo,deitke2024objaverse,chang2015shapenet}, the modeling emphasis has predominantly been placed on their \emph{appearance}, \rzz{rarely their interior structures, and certainly} not their functionalities. 

Besides these geometric and topological functional deficiencies, accuracy is another issue. For example, the majority (80\%+) of \emph{human-annotated} articulations for the storage furnitures in PartNet-Mobility~\cite{xiang2020sapien} contain errors in the motion ranges or axes which may cause part collision during articulation. We argue that all of these deficiencies and inaccuracies for existing 3D assets need to be corrected, since the main purpose of reconstructing or generating a 3D object is \emph{not} to merely look at it, but to \emph{use} it, \emph{interact with} it, perhaps to eventually \emph{manufacture} it. 

In this paper, we pose the novel problem of object \emph{functionalization}, whose goal is to convert a non-functional 3D object into one that works as intended, by fixing its
functional deficiencies through structural and geometric transformations. 
\rzz{Given the immense diversity of object functionalities, a single unified solution capable of resolving \emph{all} forms of functional deficiencies would be overly ambitious.
%
As a first step, we focus on a tractable and well-defined subset: structural functional deficiencies stemming from \emph{incomplete object structures}, as summarized in Table~\ref{tab:func_def}. Accordingly, our functionalization solution involves adding parts or connectors and adjusting part relationships, while addressing the challenge of devising a unified approach across both the dynamic and static settings.}

\begin{table}[t]
\small
\centering
\caption{A rough taxonomy of functional deficiencies related to \emph{incomplete object structures}. Dynamic deficiencies are due to missing motion-enabling elements. Static deficiencies affect usability or interactability.} 
\label{tab:func_def}
\renewcommand{\arraystretch}{1.25}
\setlength{\tabcolsep}{6pt}
\resizebox{\columnwidth}{!}{
\begin{tabular}{m{0.15\columnwidth}>{\centering\arraybackslash}m{0.28\columnwidth}p{0.45\columnwidth}}
\toprule
\textbf{Category} & \textbf{Deficiency Type} & \multicolumn{1}{c}{\textbf{Examples}} \\ \midrule

\multirow{2}{*}{Dynamic}
    & \multirow[c]{2}{=}{\centering Missing motion joints or connectors}
    & Revolute: hinges, pins, pistons\\ 
    \arrayrulecolor{lightgray}\cline{3-3}\arrayrulecolor{black}
    & & Prismatic: tracks, rails, sliders \\ 
    \arrayrulecolor{black}\cmidrule(lr){1-3}\arrayrulecolor{black}
    
\multirow{4}{*}{Static}
    & \multirow[c]{2}{=}{\centering Missing interactable parts or elements}
    & Additive: handles, knobs, pull bars \\
    \arrayrulecolor{lightgray}\cline{3-3}\arrayrulecolor{black}
    & & Subtractive: sockets, grooves, etc. \\ 
    \arrayrulecolor{darkgray}\cline{2-3}\arrayrulecolor{black}
    & Missing support elements
    & Shelves, dividers, furniture tops \\ 

\bottomrule
\end{tabular}
}
\end{table}

\rzz{To this end, we formulate object functionalization as a \emph{graph completion} problem, which offers a unified treatment of both dynamic and static deficiencies. 
Specifically, we parse an input 3D model, with part segmentation, into a new \emph{functional graph representation}, where every mesh part becomes a labeled node, 
every functional or contact relation becomes a typed edge (e.g., contact, hinge, rail, etc.), while movable nodes are further annotated with motion attributes. Under this representation, structural functional deficiencies manifest uniformly as missing nodes and/or incorrect edges, regardless of whether they are dynamic (e.g., missing hinges for revolute motion between a door and its frame and missing rails for prismatic motion) or static (e.g., a missing countertop node).}

\rzz{
We develop a neural model called a \emph{graph functionalizer} (GraFu), which takes an incomplete graph representing a non-functional object and transforms it into a complete graph representing its functional counterpart. Technically, GraFu is a graph-to-graph translation network comprising a node encoder, a Graph Transformer encoder operating on a fully-connected topology, and a DETR-style (i.e., DEtection TRansformers \cite{carion2020end}) decoder that predicts new structural nodes, functional edges (hinge, rail, attached), and motion attributes for movable parts.}

\rzz{
Graph completion with GraFu is only the first step, which is dedicated to structural (i.e., topological) functionalization without fully resolving geometric or motion precisions. In the second stage, the completed graph drives \emph{geometric realization}, where predicted connectors are instantiated and snapped into place, template meshes are inserted for new structural nodes, and the full pipeline is wrapped in a Blender add-on for interactive review and adjustment.}
A compelling side effect of such a geometric functionalization is that by grounding part motion in explicit mechanical structures rather than virtual motion axes, any errors in the initial motion annotations, as those in PartNet-Mobility~\cite{xiang2020sapien}, can be \emph{rectified} to produce physically realistic, collision-free motions; see Fig.~\ref{fig:teaser}.

\rzz{To support training GraFu and evaluation, we introduce  {\sc FurFun}-233, for Furniture Functionalization, a dataset of 233 furniture models with paired non-functional and functionalized counterparts, sourced from ShapeNet \cite{chang2015shapenet}, Objaverse \cite{deitke2023objaverse}, ABO \cite{collins2022abo}, HSSD \cite{khanna2023hssd}, and artist creations. Each model is pre-screened to ensure it contains interior structures, annotated with per-part semantic labels, and paired with a ground-truth functional graph. Functionalized counterparts include manually inserted mechanical components drawn from an expanded gallery of 9 connector types (7 hinges and 2 rails). While our formulation is general, we focus on furniture in this paper, as it represents one of the richest and most challenging categories of functional objects, encompassing both dynamic deficiencies (hinges, rails) and static ones (shelves, handles, countertops).}

\rzz{Our contributions can be summarized as follows:
\vspace{-3pt}
\begin{itemize}
\item A functional graph representation encompassing part and connector/edge labeling and motion attributes. 
\item A neural graph functionalizer for structure completion.
\item A two-stage functionalization pipeline that follows topological completion with geometry realization which can rectify predicted motion attributes via mechanical grounding.
\item A high-quality dataset {\sc FurFun}-233 to support training and evaluation of object functionalization models.
\end{itemize}
}

We evaluate our method for its motion prediction using GraFu by comparing against state-of-the-art baselines, Particulate~\cite{li2025particulatefeedforward3dobject} and SINGAPO \cite{liu2024singapo}. On the Storage Furniture subset of PartNet-Mobility (PN-M), GraFu matches their motion prediction accuracy while geometric realization substantially improves physical plausibility in terms of collision resolution and part connectivity. Our method also exhibits strong generalizability with this test since its training set does not contain any PN-M models. On HSSD, our method outperforms both baselines across all three metrics. We further demonstrate robustness via an ablation study in which input node labels are progressively \rev{masked} up to 100\%, showing that GraFu maintains competitive motion prediction and near-perfect connectivity even under severe label noise. 

\section{Related Work}
\label{sec: related work}

With rapid advances in 3D generative and spatial AI, robotics, and the pursuit of world models, the value of high-quality 3D assets has received increasing attention. Take single-view 3D reconstruction as an example, recent works such as CLAY~\cite{CLAY2024}, Trellis~\cite{Trellis2025}, and Hi3DGen~\cite{Ye2025hi3dgen} unanimously highlight the importance of \emph{high-quality} 3D training data. However, achieving this only through careful data selection and curation is insufficient. Our work seeks to bring object functionalization as a new means of quality upgrades for existing 3D assets.

Broadly speaking, object functionalization falls into the category of shape modeling and editing, a well-studied topic in computer graphics with a variety
of criteria or objectives such as symmetry~\cite{mitra_sym_2007}, stackability~\cite{li_stack_2012}, balance~\cite{prevost_stand_2013}, strength~\cite{lin_strength_2014}, foldability~\cite{li_fold_2015}, detail preservation~\cite{sorkine_meshdef_survey2008}, structure preservation~\cite{mitra_star13}, and sketch conformation~\cite{sketch_based_survey2009}, among others. ~\citet{umentani2015guided} focus on interactive design of static plank-based furniture under frictional and nail-joint constraints, offering real-time guidance to a human designer. It does not address motion-enabling components such as hinges or rails, nor does it automate the functionalization of existing 3D assets.\rev{~\citet{ureta2016interactive} proposes an algorithmic approach to insert kinematic joints, but relies on user-guided insertion signals}. To our knowledge, no prior works targeted functionalizing 3D objects via shape completion or geometric transformation.

\vspace{3pt}

\noindent\textbf{Shape Completion and Assembly.}
\am{Shape completion has been extensively studied under different settings and representations from classical hole-filling algorithms~\citep{liepa2003filling, tekumalla2004hole, ju2004robust, kazhdan2006poisson, zhao2007robust, kazhdan2013screened, centin2015poisson} to more \rev{recent deep learning approaches~\cite{yan2022shapeformer,Arora_2022_CVPR,tesema2023point, wang2022shape}.}
However, most of these techniques try to complete a portion of the shape that is missing due to occlusion or flaws in the scanning process as opposed to our technique that recovers a missing part necessary for interaction, movement, or function.}

\am{Shape assembly methods are more directly related to our approach, as they typically focus on predicting how parts and joints should be connected. For example,} JoinABLe \cite{willis_join_2022} is a learning-based assembly method which connects pairs of CAD parts by predicting the correct parametric joints, operating directly on boundary representations without class labels or human guidance. It also 
contributed the Fusion 360 Gallery assembly dataset.

\citet{adriana_fab_2014} propose a data-driven interactive system to help users design 3D models that are \emph{fabricable} by composing parts from a database of expert-created templates. It automatically aligns, parameterizes, and connects parts, with proper constraints and connectors even including screws and corner braces, so that the resulting models can be manufactured. Such a ``fabricabilization" process bears resemblance to one of our functionalization options, namely, hinge/track insertion, but without motion rectification. Generally, while fabricability and functionality are related, they are distinctive modeling goals. Operationally, our work is less about shape assembly but more on structure completion and shape transformation.

\am{In other related work \citet{lin2017recovering} reconstruct kinematic assemblies from raw 3D scans by inferring part segmentation and joint parameters, and \citet{mitra2010illustrating} analyze and visualize mechanical motion relationships. More recently, \rev{Kinematic Kitbashing~\cite{guo2025kinematic} assembles reusable parts into a prescribed kinematic graph by optimizing per-part transforms under an exemplar-based vector distance field energy aggregated across each joint's motion range.} However, all of these methods presuppose a complete assembly as input, whereas our work addresses inferring and inserting \emph{missing} functional geometry to enable articulated motion.}

\vspace{3pt}

\noindent\textbf{Motion Prediction and Generation.}
Since many object functions are realized through part motions and human-object interactions, works that predict, reconstruct, and/or generate part articulations \cite{liu_artsurvey_2025} are relevant to functionalization, but their goal is not to functionalize an object by altering its shape or part structure. Recent data-driven methods predict part articulations directly from geometry or images~\cite{liu_artsurvey_2025,li2025particulatefeedforward3dobject,liu2024singapo,le2024articulate}, but while effective at inferring \emph{where} and \emph{how} parts move, they do not address \emph{why} a motion may be physically implausible or how to correct it. Our rule-based geometric fixing stage is thus complementary: as shown in our experiments, applying hinge and rail insertion on top of predicted motion axes consistently reduces collisions, demonstrating that learned articulation and explicit mechanical grounding are mutually reinforcing. 

ArtLLM \cite{wang2026artllm} autoregressively predicts part layouts and joint structures from point clouds using a multimodal large language model, generating high-fidelity articulated assets; while it advances joint inference and part geometry synthesis, it does not address functional deficiencies in existing assets or insert physical connectors to correct implausible motions.
In this spirit, annotated \cite{xiang2020sapien} or generated (e.g., \cite{li2025particulatefeedforward3dobject,vora2025articulateobjectatop3d,qiu2025articulate}) part motions can serve to initialize motion parameters in our dynamic functionalization setting, which are then improved through functionalization. Motion prediction, and the ensuing simulation, has recently been employed for functional anomaly \emph{detection} \cite{bhunia2025interactive} by comparing observed (possibly abnormal) part articulations with anticipated (normal) ones. But the detected anomalies are not corrected.

\vspace{3pt}

\noindent\textbf{Interior Modeling.}
Object interiors are often exposed under part motion, e.g., when opening a cabinet door or drawer. As a result, recent works on geometry and interaction reconstruction such as DRAWER~\cite{xia_drawer_2025} and S2O~\cite{iliash2024s2o} contain modules which complete drawer/cabinet interiors that were hidden from the input videos or missing from the input 3D objects, 
\rz{using a category-specific template approach similar to ours.}

For representation learning, aside from volumetric representations such as NeRF \cite{mildenhall2020nerf} and 3DGS \cite{kerbl3Dgaussians}, object interiors can also be modeled by multi-slice images \cite{wang_slice3d_2024} and variants of signed distance functions (SDFs) such as 3-poled SDF \cite{chen_3psdf_2022}.
But all of these works have targeted reconstruction or generation tasks, e.g., from images or point clouds, not functionalization, for which we simply employ polygonal meshes.
Furniture interiors, along with part articulation, can be acquired in conjunction through active 3D reconstruction \cite{yan_siga23}, where a robot interacts with a physical
piece (e.g., opening/closing a drawer) while scanning the geometry.

\vspace{3pt}

\begin{figure*}
    \centering
    \includegraphics[width=\linewidth]{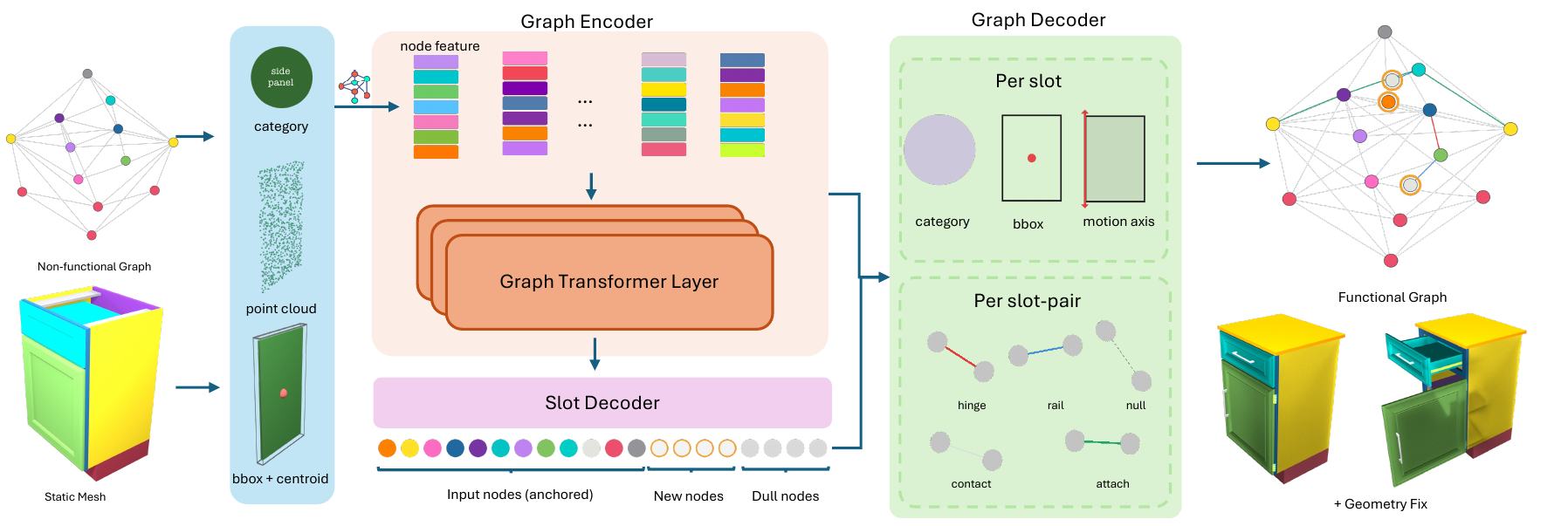}
    \caption{\textbf{Our Pipeline.} \am{Given a non-functional 3D furniture asset, the system encodes each part node using category, point cloud, and bounding box centroid features. A Graph Transformer then models inter-node relationships across the fully-connected graph. The resulting node features are passed to a DETR-style Slot Decoder, which predicts per-slot attributes (category, bounding box, motion axis) and per-slot-pair edge types (hinge, rail, contact, attach, or null) across anchored input nodes, free new-node slots, and dull (padding) slots. The output is a functionalized graph that is subsequently grounded in 3D geometry via the Geometry Fix stage, yielding a physically operable furniture model with instantiated hinge and rail mechanisms.}}
    \Description{Our Pipeline}
    \label{fig:grafu-pipeline}
\end{figure*}

\noindent\textbf{Fixing Malfunctions.}
\am{To our knowledge, the closest work to ours is \cite{hong2022fixing}, which also resolves functional deficiencies via a discrete set of fixes detected by a physics simulator. However, it focuses solely on motion malfunctions, whereas our work covers both dynamic and static deficiencies. More fundamentally, our motion rectification is not simulation-driven but a natural consequence of inserting proper mechanical connectors, which allows us to correct even the erroneous human annotations in PartNet-Mobility, which usually serves as ground truth. For static deficiencies, no prior work directly addresses the full scope we tackle: works such as \cite{hrvzica2022detection} and \cite{badamiWACV17} address the related but distinct problems of storage space detection and furniture segmentation without modifying geometry, while PhysX-Anything \cite{cao2025physx} and PhysX-3D \cite{cao2026physx} generate simulation-ready assets from scratch rather than functionalizing existing ones. Procedural tools such as the IKEA kitchen planner similarly operate under rigid predefined templates and do not reason about functional deficiencies in arbitrary 3D assets.}

\section{Method}
\label{sec:method_v2}

We address the problem of \emph{functionalization}---transforming non-functional 3D objects, particularly furniture, into physically plausible, functional counterparts.
\saura{Our approach consists of two main stages: a) The input 3D model is first converted into a graph representation that captures part-level relationships within the object. \am{In this graph, functional deficiencies, missing nodes, or incorrect edges may arise due to imperfections in the input model.
} A novel \textbf{Gra}ph \textbf{Fu}nctionalizer (\textsc{GraFu}) module then operates on this non-functional graph to predict the missing functional nodes, correct relational edges, and estimate motion axes for dynamic nodes, thereby producing a complete functional graph. b) The 3D geometry is subsequently updated by cross-referencing the predicted graph with the original input: a suite of \textit{geometric post-fix} algorithms directly embeds hinge and rail mechanisms into the 3D model for newly introduced revolute and prismatic edges, respectively, while template or bounding-box geometries are instantiated for any newly added static nodes. To support practical usability, we additionally provide a Blender add-on that allows users to interactively inspect and refine the output at each stage of the pipeline.}
\am{In the following, we discuss each of these components in detail.}
\saura{The overall pipeline of our method is illustrated in ~\Cref{fig:grafu-pipeline}.}


\subsection{Functional Graph Parsing}
\label{subsec:graph_parsing}

\saura{For a segmented 3D model, we construct a functional graph $G=(V,E)$, where $V$ denotes the set of annotated part nodes and $E$ denotes the set of inter-part contact relationships.
Each node $v \in V$ encodes three attributes: the 3D bounding box of the part, its geometric representation, and a semantic label. Part labels are assigned from a fixed vocabulary: \texttt{\{top/side/back/bottom panels, face frame, shelf, divider, bar, leg, door, drawer, handle, misc\}}. This label set covers the vast majority of structural geometries encountered in practice; parts that do not fit any named category, such as stretchers, are assigned to the \texttt{misc} label. Each edge $e \in E$ belongs to one of four types: \am{\texttt{\{contact, hinge, rail, attached\}}}. A \texttt{contact} edge encodes static structural connectivity, representing joints analogous to glued or nailed panels in physical assemblies. \texttt{Hinge} and \texttt{rail} edges capture revolute and prismatic kinematic relationships between movable parts and their fixed mounting anchors. Handles are connected to their parent door or drawer via an \texttt{attached} edge. For dynamic nodes, a motion axis $M$ is additionally defined in the local coordinate frame of the part. For revolute joints, the motion axis is selected from the bounding box edges of the part, where a positive sign denotes counter-clockwise rotation and a negative sign denotes the opposite. For prismatic joints, the motion axis is assigned as one of the canonical $\pm\{x,y,z\}$ directions.  \\
}

\subsection{GraFu Network}
\label{subsec:grafu}

\saura{\textsc{GraFu} is a graph-to-graph translation network that transforms a non-functional graph $G_{u}$ into a functional graph $G_{f}$ by predicting missing structural nodes, inferring functional edges (\texttt{hinge}, \texttt{rail}, \texttt{attached}), and estimating motion attributes for dynamic nodes. The architecture comprises three sequential components: a) a \textit{Node Encoder} that extracts per-node feature representations, b) a \textit{Graph Encoder} that models relational dependencies across nodes, and c) a \textit{Graph Decoder} that produces the functionalized output graph $G_{f}$. The complete pipeline is illustrated in \Cref{fig:grafu-pipeline}.}

\vspace{3pt}

\noindent \textbf{Node Encoder.}
\saura{The node encoder maps each node $v \in V$ in the non-functional graph $G_{u}$ to a fixed-dimensional latent embedding. The \textit{part category}, represented as a one-hot vector, is projected into a latent embedding of dimension $d_{\text{cat}} = 128$ via a lightweight MLP, capturing the semantic identity of each node. The 3D geometry of each part is encoded using a lightweight variant of PointNet++~\cite{qi2017pointnet++}, which maps the per-part point cloud to a latent embedding of dimension $d_{\text{pc}} = 64$. Additionally, the centroid of the part bounding box is encoded by a separate lightweight MLP into a latent of dimension $d_{\text{geo}} = 32$. The three embeddings are concatenated to yield a unified per-node latent representation $z_{v} \in \mathbb{R}^{224}$ for each node $v \in V$. By design, the category embedding $d_{\text{cat}}$ constitutes the dominant component of $z_{v}$, since part category is the most informative attribute for describing each node.}

\vspace{3pt}

\noindent \textbf{Graph encoder.}
\saura{To model inter-node relationships in the non-functional graph, we employ a lightweight Graph Transformer that operates over a \emph{fully-connected} graph topology, where all nodes attend to one another and absent edges are assigned \texttt{null}-type edge labels. Edge features are constructed by encoding the edge connectivity type as a one-hot vector, augmented with an 8-dimensional geometric descriptor $[\,\lVert r\rVert,\,\lVert g\rVert,\,\hat{r},\,g_{\text{axis}}\,]$ that captures the spatial relationship between part bounding boxes. Here, $\hat{r}$ denotes the signed unit vector along the centroid-to-centroid direction, and $g_{\text{axis}}\!\in\!\mathbb{R}^3_{\geq 0}$ represents the per-axis surface gap between oriented bounding boxes (set to zero where boxes overlap). The connectivity encoding and geometric descriptor are jointly projected into a 128-dimensional edge feature embedding $z_{e}$ for each node pair using an MLP. At each Graph Transformer layer, $z_{e}$ is incorporated into the attention computation as a learned edge bias:
\begin{equation}
    \text{Attn}(z^{Q}_{v},z^{K}_{v},z^{V}_{v}) = \operatorname{softmax}(\frac{z^{Q}_{v}. (z^{K}_{v})^{T}}{\sqrt{d}} + z_{e}).z^{V}_{v},
\end{equation}
where $z^{Q}_{v}$, $z^{K}_{v}$ and $z^{V}_{v}$ are Query/Key/Value derived from the node embeddings $z_{v}$. This formulation enables the model to jointly reason over both node attributes and inter-part relational structure, such as which parts are connected via hinge or rail joints, thereby improving its capacity to predict a functionally coherent output graph.}

\vspace{3pt}

\noindent\textbf{Graph Decoder.} 
\saura{To decode the learned latent representations into a functional graph, we adopt a DETR~\cite{carion2020end}-style Transformer Decoder as our Graph Decoder. Rather than using object queries as in the original DETR formulation, we introduce \textit{node slots} that serve as learned queries over graph nodes. Graph Encoder features are passed to the Graph Decoder via cross-attention. We initialize $N$ \textit{anchor slots}, one per node in the input graph, and set a global maximum slot count of $M$. To allow the decoder to predict nodes beyond those present in the non-functional input, we introduce $K$ additional \textit{free query slots} alongside the $N$ anchor slots. These free queries are tasked with recovering missing structural nodes required to make the graph functional; unused slots are zero-padded and treated as null nodes. To provide free queries with global graph context, we pool the Graph Encoder output and add its projected summary as an initialization to each free query embedding. Each decoder slot independently predicts: (i) node validity, (ii) part category, (iii) bounding box, and (iv) centroid. For every pair of nodes, the decoder additionally classifies the connecting edge into one of $\{$\texttt{rail}, \texttt{hinge}, \texttt{attached}, \texttt{contact}, \texttt{null}$\}$. When a \texttt{hinge} or \texttt{rail} edge is predicted between dynamic nodes, the motion axis can be estimated by aligning the edge on the target geometry.  The first three edge types are directed, encoding a parent-child relation between parts, while \texttt{contact} and \texttt{null} edges are treated as bidirectional.}

\vspace{3pt}

\noindent\textbf{Training.} 
\saura{To train \textsc{GraFu}, we require paired (non-functional, functional) graph data. To address the scarcity of such supervision, we introduce the FurFun-233 dataset, which provides ground-truth functional graphs. Non-functional counterparts $G_{u}$ are synthesized from each functional graph $G$ via a set of structure-corrupting augmentations, including random edge collapse, node and edge removal, node and edge label masking, and geometric flips. The resulting paired data constitutes the training corpus for \textsc{GraFu}. The whole network is trained using a combination of an anchor loss, node and edge reconstruction losses, and a graph completion loss, all computed over the synthesized training pairs. Full details of the \rev{GraFu architecture}, loss formulations, ablation studies, and augmentation strategies are provided in the supplementary material.}

\subsection{Geometric Fixing}
\label{subsec:geom_fix}

\begin{figure}
    \centering
    \includegraphics[width = \linewidth]{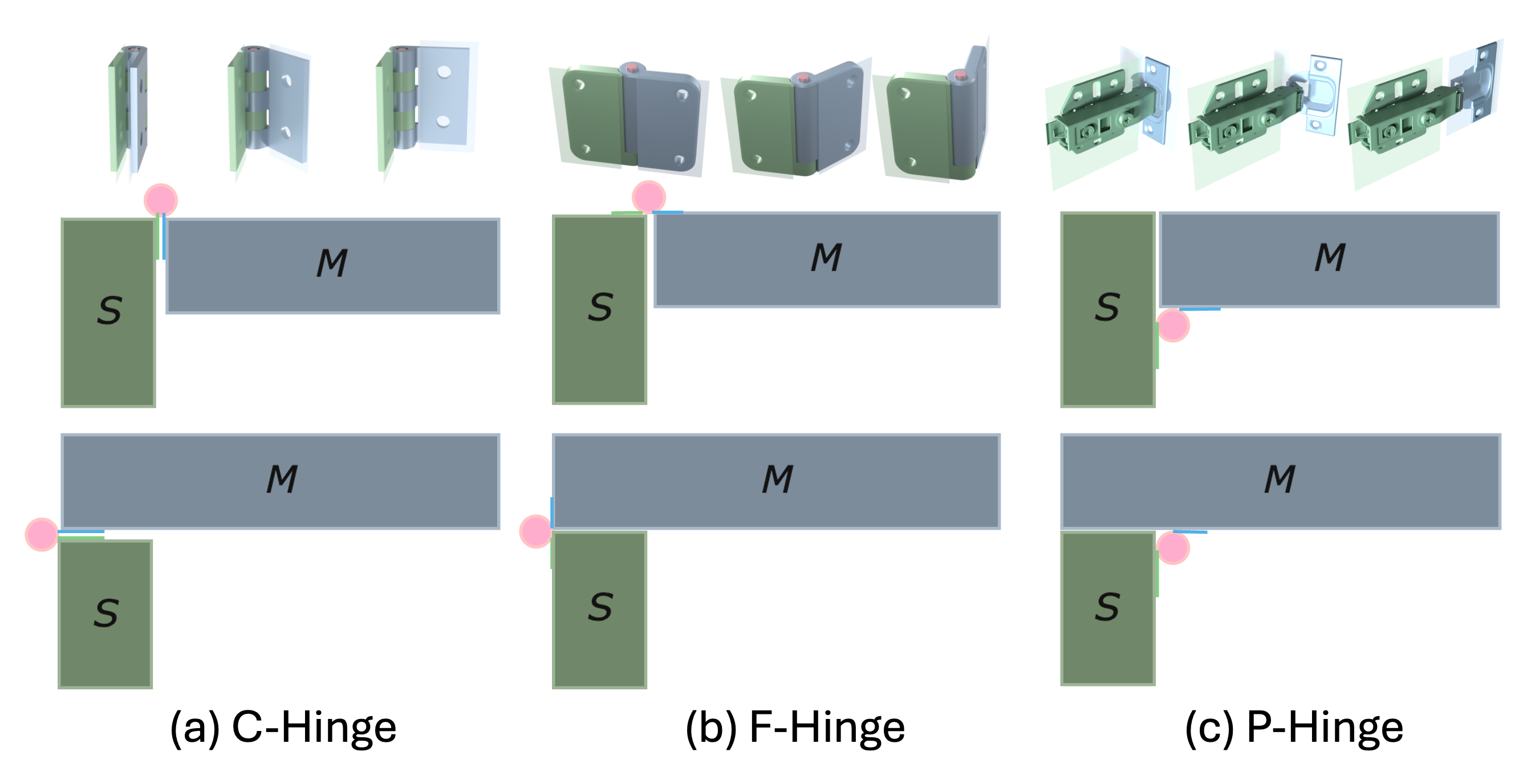}
    \caption{Various application conditions for Contact-face snapping, Flushing-face snapping, and Perpendicular-face snapping hinges under counter-clockwise rotation motion.}
    \label{fig:hinge_config}
\end{figure}

\am{Given the predicted graph and input mesh, Stage~2 realizes the prediction in 3D via dynamic and static geometric fixes. For every predicted hinge or rail edge, the corresponding mechanical geometry is instantiated and integrated into the mesh, with dynamic nodes inheriting the embedded motion animations — mirroring how real-life furniture motion is governed by physical joints rather than virtual axes. For every predicted new node, a template mesh is instantiated and surfaced through a Blender API for the user to visualize, tune, accept, or deny.}


\vspace{3pt}

\noindent\textbf{Hinge Insertion.}
\am{Inserting actual hinge geometries into the static mesh offers multiple benefits. First, it ensures physical motion plausibility by keeping all parts connected throughout motion, with the range naturally constrained by the hinge. Second, the hinge carries its own local revolution system, snapping it between the movable and static parts rectifies the predicted motion axis, making the revolute motion geometrically grounded and physically consistent.}


\am{Physically plausible revolute motion requires \emph{bilateral surface contact}. This means that the joint must simultaneously contact both the static frame and movable part at geometrically compatible locations. We identify three hinge types based on surface arrangement (Fig.~\ref{fig:hinge_config}): C-hinge for two contacting faces, F-hinge for flushing faces, and P-hinge for perpendicular interior faces. Together these three types empirically cover all furniture revolute joints, with C-hinge being the most prevalent in practice.}


\am{For hinges, we apply a template-based snapping algorithm using a gallery of the three hinge types, each annotated with static/dynamic snap planes and a rotation axis. Given a predicted hinge edge and its two nodes, we detect the presence of contacting, flushing, or interior perpendicular faces to determine the appropriate hinge type, after which a local geometric solver finds the optimal hinge placement.}


Given a static-side part $S$ and a dynamic-side part $D$ from the predicted edge, we (i) align the rotation axis to the network-predicted \texttt{hinge\_axis\_signed} on $D$, (ii) place the hinge along the predicted \texttt{hinge\_border} face of $D$, and (iii) snap it to the actual surfaces of $S$ and $D$ by solving a $3\!\times\!3$ linear system for a translation $\Delta$:
\[
\begin{aligned}
\mathbf{n}_s\!\cdot\!\Delta &= t_s, \quad
\mathbf{n}_d\!\cdot\!\Delta = t_d, \quad
\hat{\mathbf{a}}\!\cdot\!\Delta = 0,
\end{aligned}
\]
where $(\mathbf{n}_s,\mathbf{n}_d)$ are the world-space snap-plane normals, $(t_s,t_d)$ are robust median ray-cast distances from the snap planes to $S$ and $D$ along $-\mathbf{n}$ and $\hat{\mathbf{a}}$ is the rotation-axis direction.

\vspace{3pt}

\begin{table*}[t]
\centering
\caption{Quantitative evaluation on PN-M (345 models, ``zero-shot'') and HSSD (50 models). \emph{GT-aligned} matches dataset annotations one-to-one; \emph{Reasonable} is human-verified plausibility (PN-M only). Collision and connectivity cells report \textbf{raw} (\textbf{$+$\textsc{GeoFix}}), where the second value in brackets is obtained by applying our dynamic connector insertion on top of the method's predicted motion axes for motion rectification. The \emph{GT annotation} row applies \textsc{GeoFix} on top of ground-truth motion axes annotated by humans, isolating the contribution of geometric realization. \emph{Ours} produces a single integrated result, so no separate raw/$+$\textsc{GeoFix} split is shown. Best per slot in \textbf{bold}, second best \underline{underlined} (\emph{GT annotation} excluded from ranking).}
\label{tab:quant}
\setlength{\tabcolsep}{4pt}
\renewcommand{\arraystretch}{1.15}
\resizebox{0.85\linewidth}{!}{%
\begin{tabular}{l cc cc | c cc}
\toprule
& \multicolumn{4}{c}{PartNet-Mobility}
& \multicolumn{3}{c}{HSSD} \\
\cmidrule(lr){2-5} \cmidrule(lr){6-8}
& \multicolumn{2}{c}{Motion Correctness $\uparrow$}
& Collision $\downarrow$ & Connectivity $\uparrow$
& Motion $\uparrow$
& Collision $\downarrow$ & Connectivity $\uparrow$ \\
\cmidrule(lr){2-3}
Method
& GT-aligned & Reasonable
& (\%) & (\%)
& GT-aligned
& (\%) & (\%) \\
\midrule
Particulate~\cite{li2025particulatefeedforward3dobject}
& \underline{87.6\%} & \textbf{96.5\%}
& \underline{7.9} (5.5) & 13.2 (96.5)
& \rev{27.4\%}
& 29.0 (28.7) & \underline{62.8} (58.0) \\
SINGAPO~\cite{liu2024singapo}
& \textbf{87.8\%} & 93\%
& 28.7 (27.0) & \underline{48.4} (73.8)
& \underline{59.4\%}
& \underline{28.7} (24.6) & 51.7 (98.1) \\
\emph{GT annotation}
& -- & --
& 19.2 (5.2) & 49.1 (97.2)
& --
& 25.0 (25.0) & 82.1 (97.7) \\
Ours
& 87.0\% & \underline{96.3\%}
& \textbf{4.4}\phantom{\,/\,99.9} & \textbf{98.1}\phantom{\,/\,99.9}
& \textbf{85.8\%}
& \textbf{15.1}\phantom{\,/\,99.9} & \textbf{98.6}\phantom{\,/\,99.9} \\
\bottomrule
\end{tabular}}%
\end{table*}
\vspace{3pt}

\noindent\textbf{Rail Insertion.}
\am{Rail insertion follows a similar template-based approach, with each template pre-annotated with static and dynamic snap planes. The rail is oriented to the predicted sliding axis, snapped to the movable part's main body, and scaled to its length. Since most 3D assets lack mechanical compatibility, gaps between the rail and mounting panel are bridged with an additional supporting block. The inserted rail constrains motion range proportionally to the part size and ensures the movable part remains physically connected throughout animation.}

\vspace{3pt}
\noindent\textbf{\rev{Static Functionalization.}} In addition to functional edge instantiation, our GraFu also predicts new nodes that complement the model structure, such as additional handles, top panel (which is the most commonly missed item in the PartNet-Mobility dataset), shelves and dividers. We develop respective algorithms for each of the categories, detailed in the Supplementary Materials.

\vspace{3pt}
\noindent\textbf{Interactive Correction.}
\rev{Upon functional graph prediction, all dynamic functional connectors, top panels and handles are automatically instantiated from the predicted motions and newly added nodes. Human intervention contributes what automation cannot decide, namely personal preference and design aesthetics beyond the automatic installation: template variants, handle styles and positions, additive or subtractive forms, and the interior layout of shelves and dividers. Our Blender API bridges this window, letting users express design intent and further tune or correct functionalized models after automation.}



\section{Experiments}
\label{sec:exp}

We evaluate our functionalization output along three axes: \emph{motion correctness}, \emph{motion plausibility}, and \emph{structural completeness}. Motion correctness measures the position and sign of the predicted motion axis. Motion plausibility captures the collision rate and rootedness of the dynamic geometry---whether movable parts detach from or penetrate the static body during articulation. Structural completeness measures whether all required structural elements are present.

\vspace{3pt}

\noindent\textbf{FurFun-233 Dataset.}
\am{To facilitate further investigations on \emph{functionalization}, we introduce {\sc FurFun}-233, a dataset of \textbf{233} furniture assets with paired \emph{non-functional} and \emph{functionalized} versions. Assets are sourced from ShapeNet~\cite{chang2015shapenet}, Objaverse~\cite{deitke2023objaverse}, ABO~\cite{collins2022abo}, HSSD~\cite{khanna2023hssd}, and artist creations, all pre-screened to ensure non-monolithic interiors. Functionalized versions are produced by human annotators via per-part labeling and manual insertion of mechanical components (\textit{e.g.}, handles, hinges, and rails), with appropriate scaling and placement to match the original design intent, and are coupled with ground truth functional graphs.}

\vspace{3pt}

\noindent\textbf{Datasets and Training.}
We benchmark on two datasets: the Storage Furniture subset of PartNet-Mobility (PN-M), and 50 randomly sampled models from HSSD. For the PN-M benchmark we train a separate checkpoint that \emph{excludes} all 33 ShapeNet-sourced models to avoid data leakage; for the remaining experiments we train on the FurFun training split (80\%). We set $M{=}55$, $K{=}4$, and use Adam ($lr{=}3\mathrm{e}{-}4$) with cosine annealing. Training data is augmented by extracting non-functional graphs from functional ones (see Supplement). Training completes in 1.5 hours on a single RTX~4090.

\subsection{Quantitative Results}

\noindent\textbf{Baselines and Protocol.}
We compare against Particulate~\cite{li2025particulatefeedforward3dobject} and SINGAPO~\cite{liu2024singapo}, as well as ground-truth annotations, on PN-M and HSSD. PN-M segmentations and category labels are mapped to our taxonomy via keyword matching (\eg \texttt{panel side} maps to the side panel category), covering 67\% of the parts, \rev{followed by post-verifications to correct any mislabelings}; the remainder of PN-M (\eg ambiguous \texttt{vertical bar} parts) and the HSSD test set are manually segmented at the part level. The input non-functional graph uses contact-only edges, determined by bounding-box intersection between parts. 

To ensure a fair comparison for the two baselines, we expose both of them to maximal GT information: for Particulate, we replace its segmented meshes with GT part meshes after shape matching; for SINGAPO, we supply the GT mobility graph (bypassing its LLM-based prediction), inpaint its bounding-box prediction with the GT bboxes during diffusion, and provide GT meshes upon retrieval. 

\vspace{3pt}

\noindent\textbf{Generalizability.}
Note that both baselines were previously trained on PN-M, whereas our method is ``zero-shot" on this dataset, since our training data does not contain any ShapeNet, and hence PN-M, models. This suggests that a comparison on PN-M is also a test of generalizability for our method.

\vspace{3pt}

\noindent\textbf{Metrics.}
Because PN-M contains substantial model duplication with differing motion annotations, we report two motion-correctness measures: \emph{GT-aligned}, requiring a one-to-one match with the PN-M annotation, and \emph{Reasonable}, accepting any motion observed on other GT annotations of the same geometry. For collision, we densely sample 50 frames spanning each part's rest-to-full-extent motion and count frames in which the movable and static parts intersect; the per-object rate is averaged over all movable parts. Connectivity is measured by the distance between the movable and static parts along the motion: if this distance exceeds 5\,mm (models are normalized to a unit cube), the part is deemed detached, and we report the fraction of movable parts that remain connected. 

\vspace{3pt}

\noindent\textbf{Results.}
\Cref{tab:quant} summarizes the quantitative comparison results. GraFu predicts plausible motion for movable parts, and mechanical-part insertion substantially improves physical motion rationality. The geometric fix is method-agnostic: applied on top of Particulate or SINGAPO motion axes, or human-annotated GT, it consistently improves both collision and connectivity. On PN-M, our method matches the baselines on motion prediction accuracy despite being zero-shot, indicating strong generalizability across datasets. 

On HSSD, our method outperforms both baselines across all metrics, while it is worth mentioning that our training data does contain models from HSSD, while the two baselines did not. Particulate tends to predict motion axes that detach the movable part from the static frame, yielding a low collision rate but also low connectivity---our mechanical-part insertion repairs both failure modes.

\vspace{3 pt}

\noindent\textbf{Robustness to \rev{Missing Labels}.}
We further ablate on partially labeled inputs, simulating the case where some part labels are missing or unknown. We mask 10\%, 30\%, 50\%, and 100\% of input node labels at inference (\Cref{tab:corruption_ablation}). \rev{The performance degrades monotonically with corruption, while remaining in a reasonable band. The quantitative numbers are reported from a five-run average to account for the stochasticity in random node masking.}

\vspace{3 pt}

\noindent\textbf{{\rev{Static Functionalization.}}} \rev{We further evaluate the structural completeness enabled by our static functionalization on the PN-M dataset. Through GraFu prediction and the automatic geometric fix, 100\% of the top-less models are completed with a top panel, and 95.4\% of the models are paired with graspable handles. Interior completion, invoked upon user request, applies to models containing compartments that can accommodate internal structures; among these, 98.5\% are successfully populated with shelves and dividers. We provide further qualitative results, detailed statistics and failure case analysis in the supplementary material.}

\begin{table}[t]
\centering
\caption{Robustness ablation on PN-M and HSSD under varying levels of \rev{missing node labels}. At \rev{masking} rate $r$, a random fraction $r$ of input node material labels is replaced with \texttt{unknown}.}
\label{tab:corruption_ablation}
\setlength{\tabcolsep}{4pt}
\renewcommand{\arraystretch}{1.1}
\resizebox{\linewidth}{!}{%
\begin{tabular}{c c c c | c c c}
\toprule
& \multicolumn{3}{c}{PartNet-Mobility}
& \multicolumn{3}{c}{HSSD} \\
\cmidrule(lr){2-4}\cmidrule(lr){5-7}
\rev{Masking}
& Motion Corr. $\uparrow$ & \rev{Coll.} $\downarrow$ & \rev{Conn.} $\uparrow$
& Motion Corr. $\uparrow$ & \rev{Coll.} $\downarrow$ & \rev{Conn.} $\uparrow$ \\
rate
& (GT-aligned) & Rate (\%) & (\%)
& (GT-aligned) & Rate (\%) & (\%) \\
\midrule
10\%  & \rev{86.23\%} & \rev{4.49} & \rev{98.28}\% & \rev{72.68\%} & \rev{14.71} & \rev{98.72}\% \\
30\%  & \rev{85.76\%} & \rev{4.36} & \rev{97.88}\% & \rev{70.33\%} & \rev{16.02} & \rev{98.53}\% \\
50\%  & \rev{85.44\%} & \rev{4.46} & \rev{97.83}\% & \rev{62.68\%} & \rev{16.82} & \rev{96.93}\% \\
100\% & 82.07\% & 5.3 & 96.79\% & 52.87\% & 21.8 & 96.10\% \\
\bottomrule
\end{tabular}}
\vspace{-1 em}
\end{table}

\subsection{Qualitative Results}

\begin{figure*}
    \centering
    \includegraphics[width=0.82\linewidth]{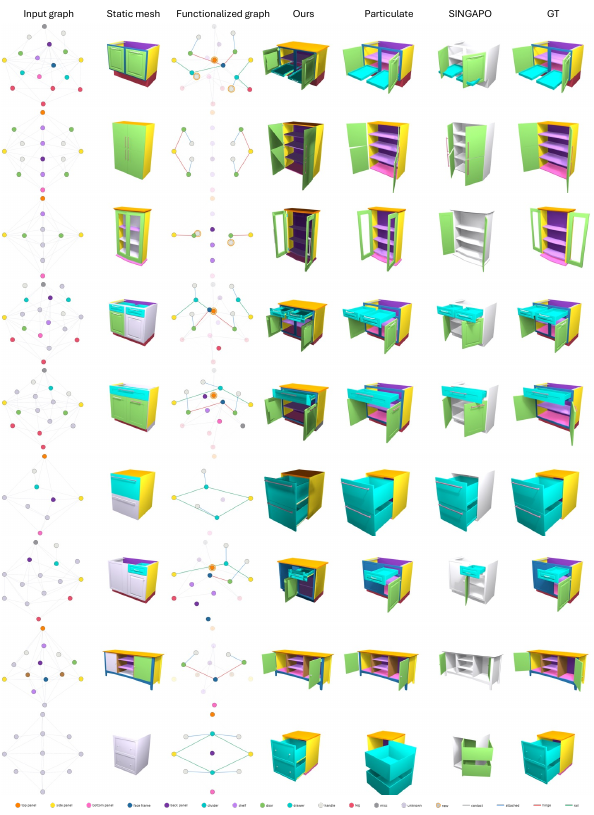}
    \caption{Qualitative results and comparisons to Particulate and SINGAPO on test samples from PN-M (top 7 rows) and HSSD (bottom 2 rows). For our results, we mainly show GraFu graph completion with added nodes and edges highlighted in the functionalized graphs (third column). From top-down, the input graphs (left column) are increasingly ``corrupted" with less nodes semantically labeled, shown as grey vs.~colored nodes. \rev{All results are obtained from automatic functionalization executions.}}
    \label{fig:result grid}
\end{figure*}

\begin{figure*}
    \centering
    \includegraphics[width=\linewidth]{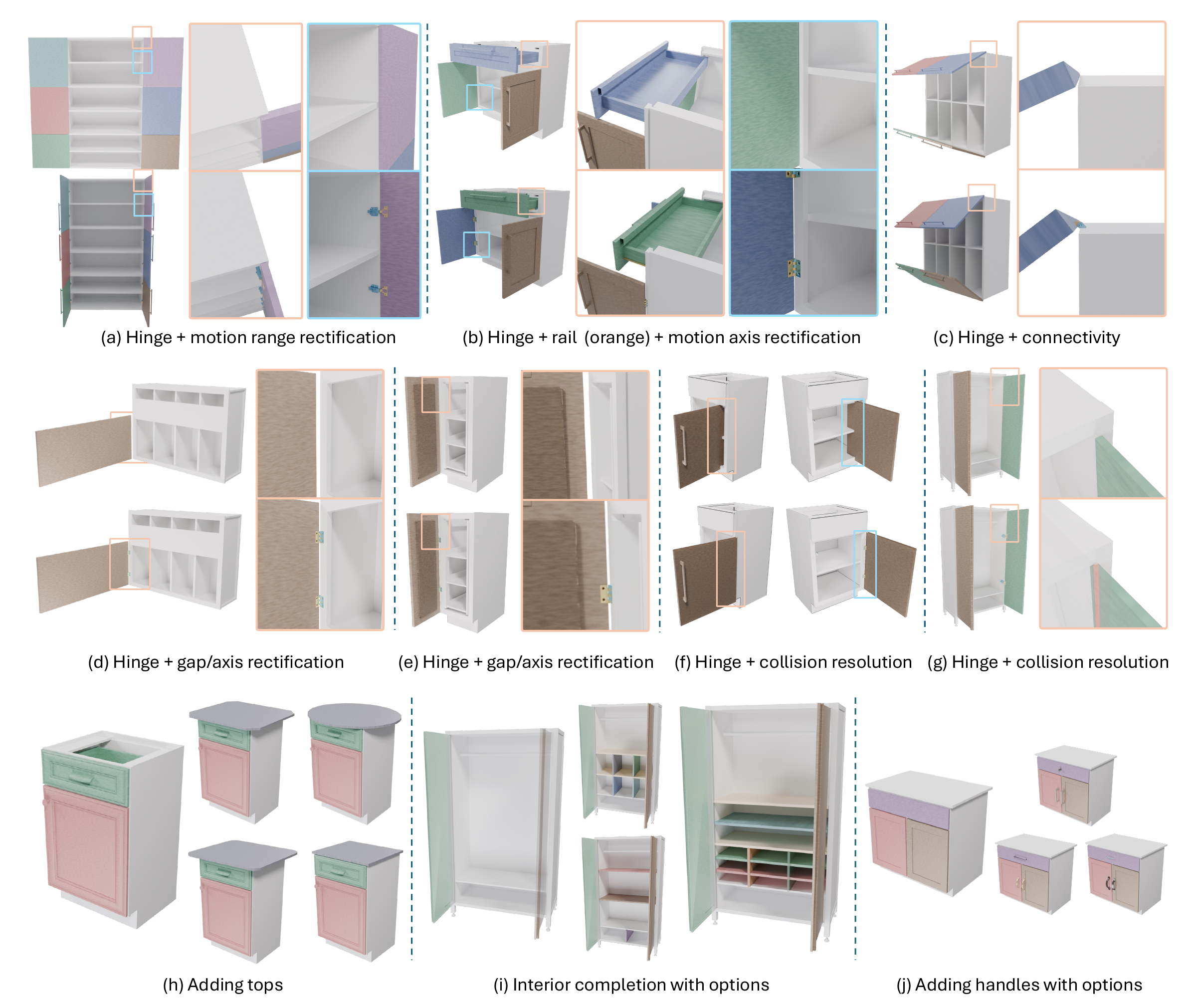}
    \caption{A gallery of qualitative results from our object functionalization tool. Dynamic functionalization is shown in (a-g), \rev{with motion axis read from GT annotations and geometric fix automatically executed.} The top is before functionalization and bottom is after. We highlight the fixes in both zoom-ins and sub-figure captions. \rev{Static functionalizations (h-j) with user-in-the-loop design are shown in the bottom row. Users can alter design styles of top panel, interior, and handles through the Blender Add-On.}}
    \label{fig:lots_of_results}
\end{figure*}

\begin{figure*}
    \centering
    \includegraphics[width=\linewidth]{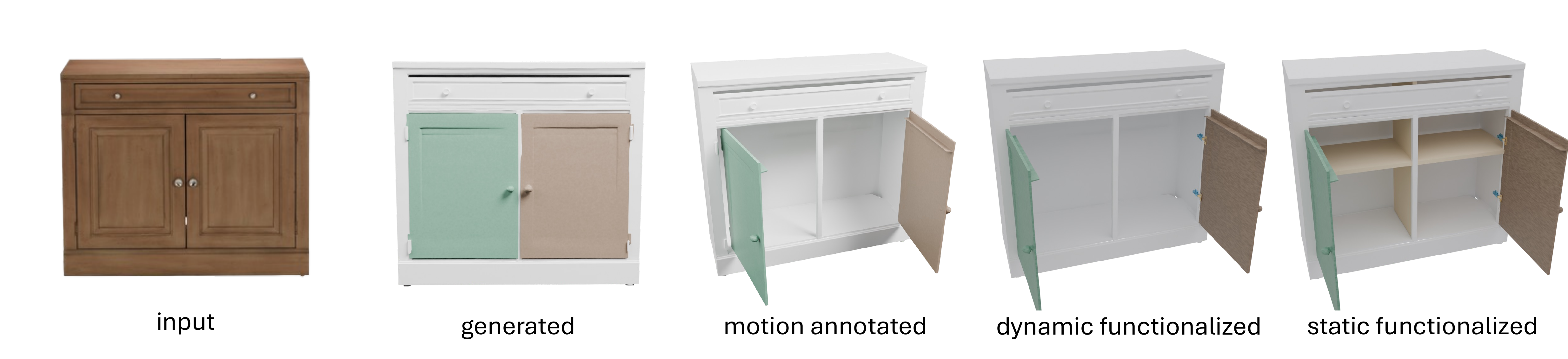}
    \caption{Dynamic and static functionalizations applied to a 3D model obtained from single-view 3D reconstruction by Omniparts \cite{yang2025omnipart}. \rev{The motion axes are read from human annotations, with automatically executed dynamic functionalization and user-designed interiors.}}
    \label{fig:omni}
\end{figure*}

\Cref{fig:result grid} presents qualitative outputs of GraFu and its geometric refinement. From top to bottom, input graph nodes are progressively \rev{masked} (\rev{masked} nodes shown in pale gray); the bottom two rows are fully unlabeled. Our method robustly emits functional edges and produces realistic articulation through mechanical-part insertion. Newly introduced nodes are highlighted: rows 1, 4, 5, and 7 gain a top panel; rows 1 and 3 gain handles, all installed automatically by our functionalization pipeline. \Cref{fig:lots_of_results,fig:omni} further demonstrate our functionalization operations in inserting mechanical parts, bridging connectivity, resolving collisions, structural completion and application to generated objects.

\vspace{3pt}

\noindent\textbf{Comparison to LLM-based Methods.}
Finally, we compare against an LLM-based baseline that uses a commercial LLM to complete non-functional graphs and BlenderMCP to perform mechanical-part insertion. Qualitative results are provided in Supplementary.

\section{Conclusion, Limitations, and Future Work}
\label{sec: conclusion}

We present object functionalization, a novel 3D modeling task to convert visually plausible but non-functional 3D assets into objects that operate as intended. To our knowledge, this is the first systematic attempt to correct major functional deficiencies in both dynamic and static settings. Our functionalization pipeline operates in two stages, by disentangling relation/topology prediction, via a graph completion neural network (GraFu), and geometry realization driven by the predicted graph which can help refine the predicted motion parameters (e.g., for collision avoidance).

\begin{figure}
    \centering
    \includegraphics[width=\linewidth]{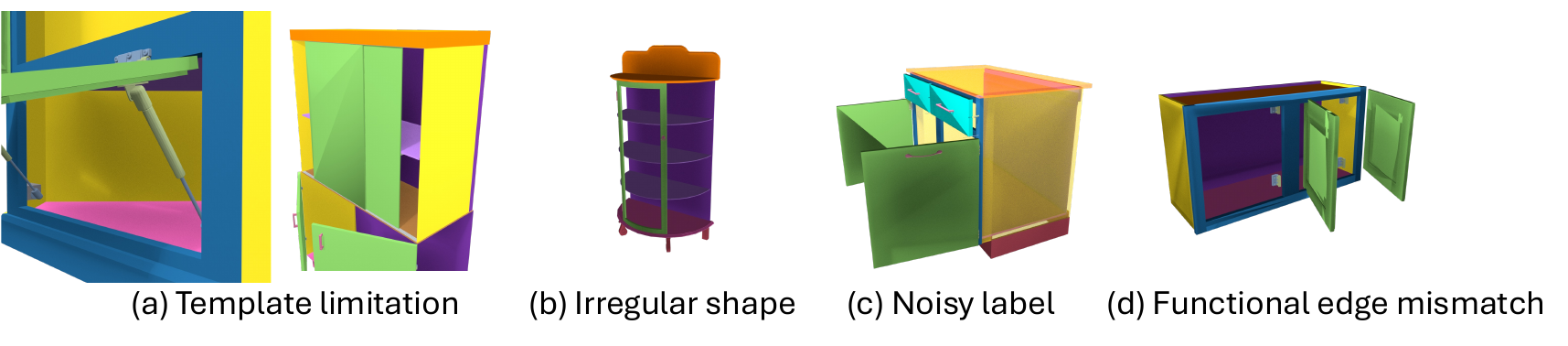}
    \caption{Limitations and failure cases. Our method is bounded by the available template gallery, which does not yet cover piston or sliding-door motions (a); struggles with highly irregular shapes that violate the bounding-box assumptions of our geometric solver (b); inherits errors from contaminated input labels (c); and is susceptible to cascaded failures when GraFu predicts the wrong mounting node for a functional edge (d)}

    \label{fig:limitation}
\end{figure}

While effective, our approach is still limited on several fronts. First, it can only functionalize parts that have been segmented, which would not cover, for example, the top drawer in Figure \ref{fig:omni}. This example also reveals that our current method can only add whole parts, not partial structures, such as the missing panels of a drawer. Second, our method does not handle geometric deficiencies that compromise physical validity at the modeling or mesh level (e.g., multiple doors in one segment or shell-only models). Last but not least, our current focuses has been on a limited set of motion types (e.g., no chained motions) and a narrow set of object categories (i.e., furniture) as dictated by the training set we constructed. Figure \ref{fig:limitation} shows several failure cases that reveal additional limitations.

Despite these limitations, we believe that our first attempt at addressing such an important modeling problem opens the door for further research. Enriching our {\sc FurFun} dataset and extending our framework to accommodate broader functional mechanisms and additional object categories would increase applicability. Beyond furniture, functionalization has potential applications in fabrication, simulation, and embodied AI, where functional correctness is critical. We hope this work encourages the community to move beyond appearance-driven 3D assets and toward models that are not only visually plausible but also physically usable and functional.

\begin{acks}
We sincerely thank all anonymous reviewers for their careful reading and constructive feedback. We also thank Dr.~Francisca Gil-Ureta and Aditya Ganeshan for the insightful discussion. This project is funded by NSERC Discover Grant RGPIN-2026-07641. 
\end{acks}

\thispagestyle{empty} 
\twocolumn[
{\SMtitlefont Functionalization via Structure Completion and Motion Rectification \par}
\vspace{0.75em}
{\SMsubtitlefont (Supplementary Material)\par}
\vspace{1.em}
]
\appendix


\begin{revblock}
\section{GraFu Architecture}
\label{sec: grafu architecture}
This section details the GraFu architecture. GraFu is composed of three modules: a node encoder, a graph encoder, and a graph decoder. Conceptually, the node encoder converts the per-part information, including category label, bounding box with centroid, and surface point cloud, into a compact latent. The graph encoder then relates these latents to one another with respect to the input graph topology. Finally, the graph decoder, conditioned
on the encoded graph, decodes the final functional graph. Each module is detailed below.

\subsection{Node Encoder}
\begin{figure}
    \centering
    \includegraphics[width=\linewidth]{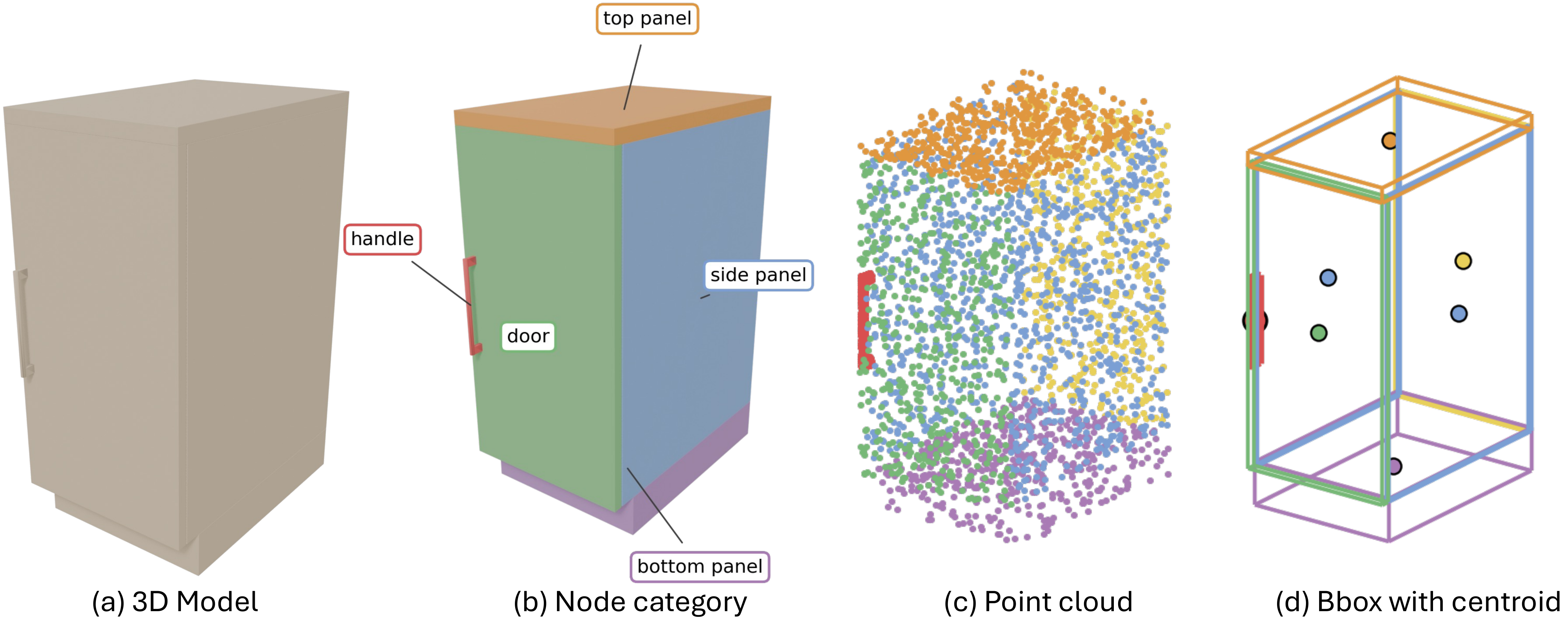}
    \caption{The node attributes encoded by GraFu. For (a) an input model, its (b) per-part category, (c) surface point cloud and its (d) bounding box and centroids, are encoded and concatenated to form a node feature.}
    \label{fig:supp_node_input}
\end{figure}
\Cref{fig:supp_node_input} illustrates the three node attributes encoded by the node encoder. Each node represents a geometric part segmented out from the input mesh, the associated attributes are defined as follows:
\begin{itemize}
    \item \textbf{Category}, from the preset list of 13 part labels plus an
          \texttt{unknown} type, $n_{\mathrm{cat}} \in
          \{0,1\}^{14}$. The \texttt{unknown} type absorbs unlabeled parts, both
          from label-masking augmentation and from missing labels at test
          time.
 \item \textbf{Bounding box and centroid.}
      The axis-aligned bounding box and centroid of the part, which together
      capture its extent and position:
      $n_{\mathrm{geo}} = [\, n_{\mathrm{bbox}} \,\|\, n_{\mathrm{centroid}} \,]
      \in \mathbb{R}^{9}$, where
      $n_{\mathrm{bbox}} := (\mathit{bbox}_{\min}, \mathit{bbox}_{\max})
      \in \mathbb{R}^{6}$ stores the two corner coordinates and
      $n_{\mathrm{centroid}} \in \mathbb{R}^{3}$ the part's centroid.
      These give the network a coarse notion of part size and placement.
    \item \textbf{Point cloud}, 512 points sampled from the part surface to
          provide geometric cues, $n_{\mathrm{pc}} \in
          \mathbb{R}^{512\times3}$.
\end{itemize}

$n_{\mathrm{cat}}$ is encoded by a linear projection into
$\mathbb{R}^{128}$; $n_{\mathrm{geo}}$ by a two-layer MLP into
$\mathbb{R}^{32}$; and $n_{\mathrm{pc}}$ by a lightweight two-stage
PointNet++~\cite{qi2017pointnet++} into $\mathbb{R}^{64}$. The three embeddings
are concatenated into the node latent $z_v \in \mathbb{R}^{224}$. 

\subsection{Graph Encoder}
\begin{figure}
    \centering
    \includegraphics[width=\linewidth]{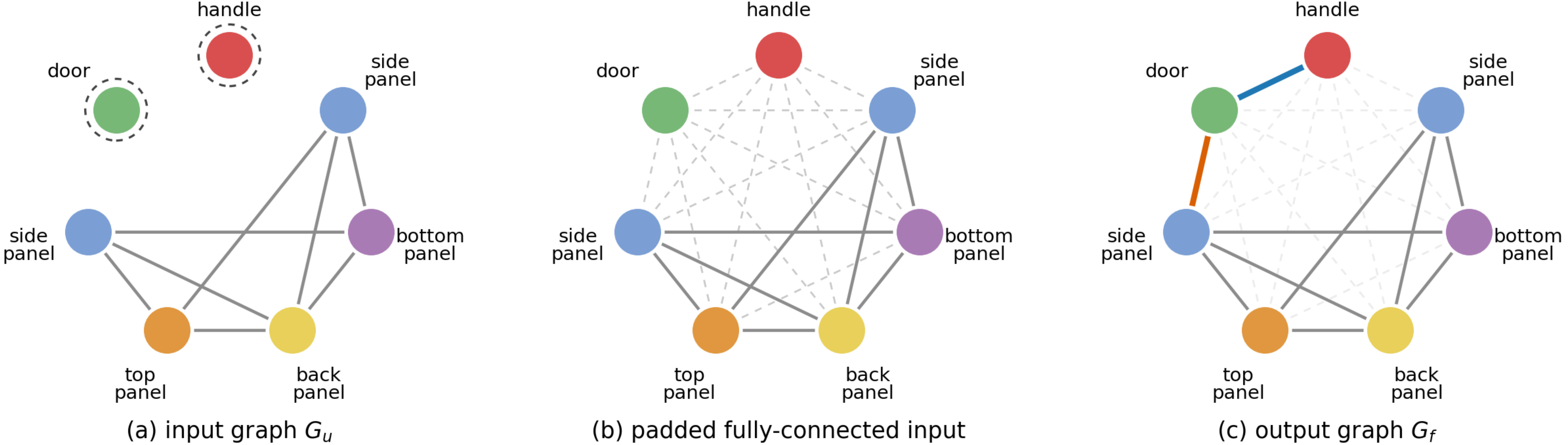}
    \caption{Edge padding mechanism on input graphs.}
    \label{fig:fully connected}
\end{figure}

\begin{figure}
    \centering
    \includegraphics[width=\linewidth]{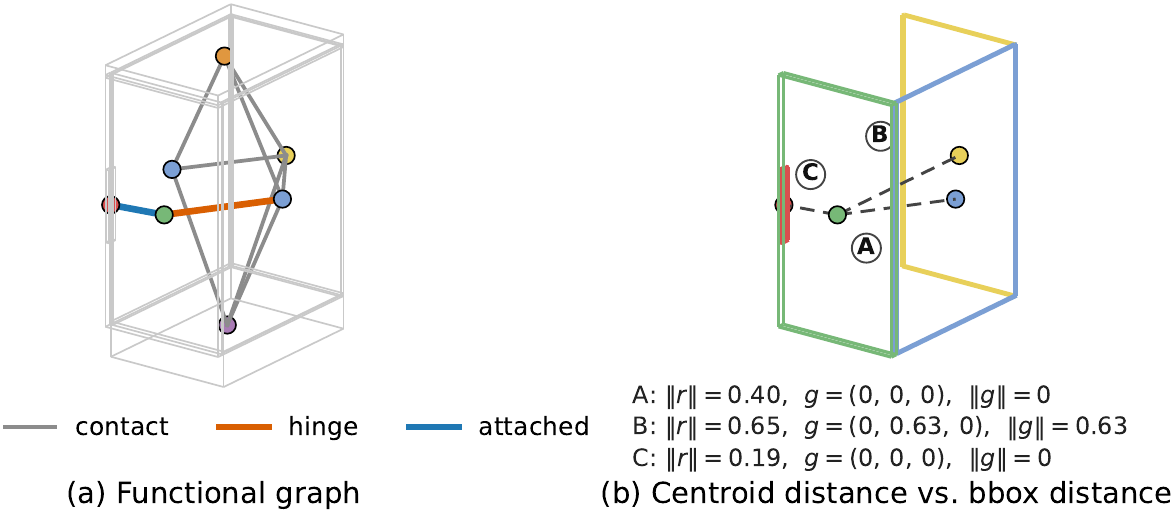}
    \caption{Bounding box distance reflects faithful contact distance between nodes, providing complementary spatial awareness besides Euclidean centroid distance.}
    \label{fig:edge features}
\end{figure}

The graph encoder allows the encoded node features to attend to one another with respect to the graph topology, yielding a relation-aware graph encoding. With node encoding described above, the design choices here concern how the inter-node relations (i.e., edges) are represented and learned. Each node pair is described by the following attributes:
\begin{itemize}
    \item \textbf{Edge type.} $e_{cat}\in \{0,1\}^5$. One of five types in $[$\texttt{contact},
          \texttt{hinge}, \texttt{rail}, \texttt{attach},
          \texttt{null}$]$, represented as a one-hot vector. 
    \item \textbf{Node distance.} $r \in \mathbb{R}^3$. The Euclidean distance between the two centroids, together with the signed unit direction. 
    \item \textbf{Bounding-box distance.} $g \in \mathbb{R}_{0+}^3$. The distance between the two nodes' bounding boxes along each axis, lower-bounded by zero under overlap.
\end{itemize}

We now elaborate on two design choices introduced above. First, the \texttt{null} edge: we pad the input graph to a fully connected graph by assigning \texttt{null} edges to the empty edge slots during graph encoding. This allows new functional edges to form on newly added nodes or disjoint node pairs at input. Edge creation reduces to a \emph{reclassification} of \texttt{null} edge types rather than the insertion of new edge entries at inference time. An example is shown in \Cref{fig:fully connected}, where the two isolated nodes (a) at input time are padded with null edges (b), and later reclassified to functional edges upon inference (c). 

Second, the bounding-box distance, which further consolidates spatial awareness between nodes. An example is shown in \Cref{fig:edge features}, which shows the pairwise distance between the door (green) towards the side panel (blue), the back panel (yellow) and the handle (red). The centroid distance gives non-zero measurements for all three pairs, though the door is indeed contacting two of them at the rest state. Using the bbox distance reflects the correct distance rankings among these three pairs.

\begin{figure}
    \centering
    \includegraphics[width=\linewidth]{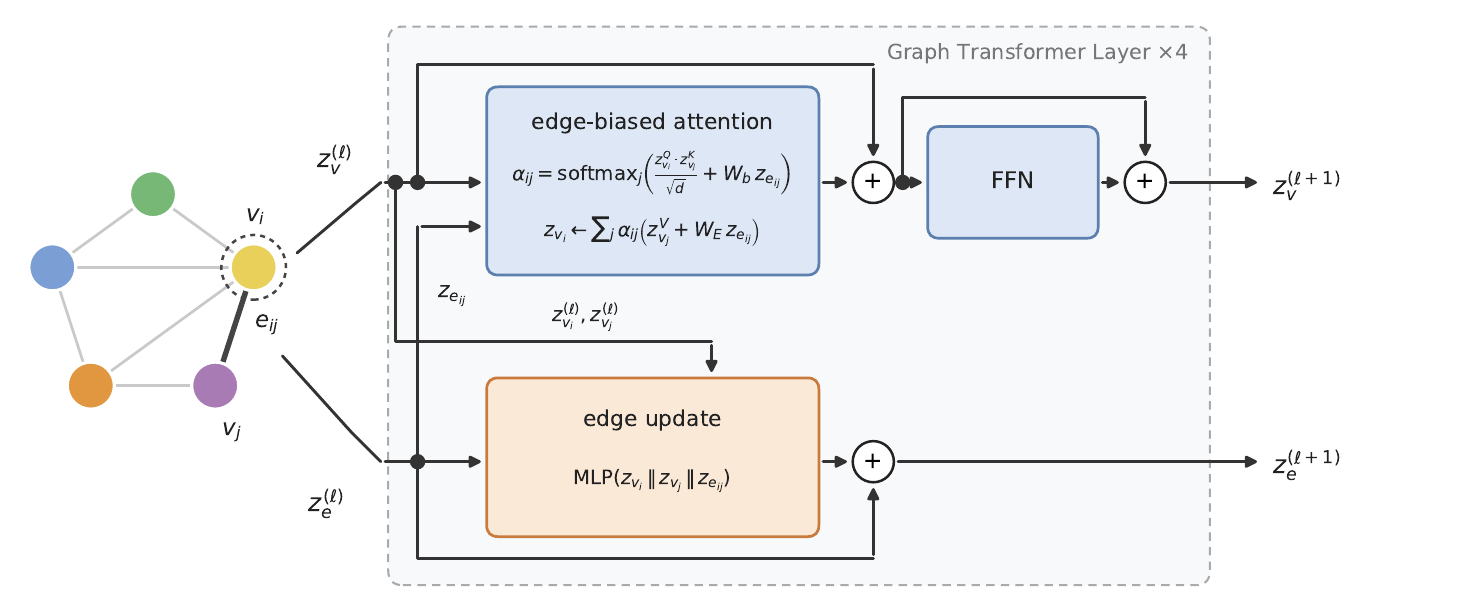}
    \caption{Graph encoder architecture. It takes 4 cascaded graph transformer layers that cross-attend each pair of nodes based on their edge connection.}
    \label{fig:supp_gt_layer}
\end{figure}

The edge type is embedded, the geometric term $[\,\|r\|, \|g\|, \hat r, g\,]$ is encoded by a small MLP, and their concatenation is projected to a 128-dimensional edge feature $z_e$. Together with the node features, $z_e$ enters a stack of four Graph Transformer layers (\Cref{fig:supp_gt_layer}). Each layer is a pre-norm transformer block over all node pairs, made edge-aware at two points: $z_e$ contributes a per-head bias $W_b z_e$ to the attention score and a term $W_E z_e$ to the attended value, so an edge steers both how strongly its endpoints attend to each other and what content they exchange. Each layer then re-encodes every edge from its two endpoint features, so node and edge representations refine each other layer by layer.

\subsection{Graph Decoder}

The graph decoder is a DETR-style~\cite{carion2020end} transformer decoder that decodes a fixed budget of $M{=}55$ slots into the output graph through four layers of slot self-attention and cross-attention to the graph encoding. The slots take three roles:
\begin{itemize}
    \item \textbf{Anchored slots} ($N$, one per input node), initialized
          from the corresponding encoder feature; they always decode to a
          real node, since functionalization never removes input parts.
    \item \textbf{Free slots} ($K{=}4$), learned queries conditioned on a
          pooled summary of the graph encoding; they instantiate the
          missing parts.
    \item \textbf{Padding slots} ($M{-}N{-}K$), supervised to be
          non-existent.
\end{itemize}

The inference goal of the decoder is to, first, predict the node attributes for each slot, and second, predict the edge relations between predicted nodes. 

\begin{figure}
    \centering
    \includegraphics[width=\linewidth]{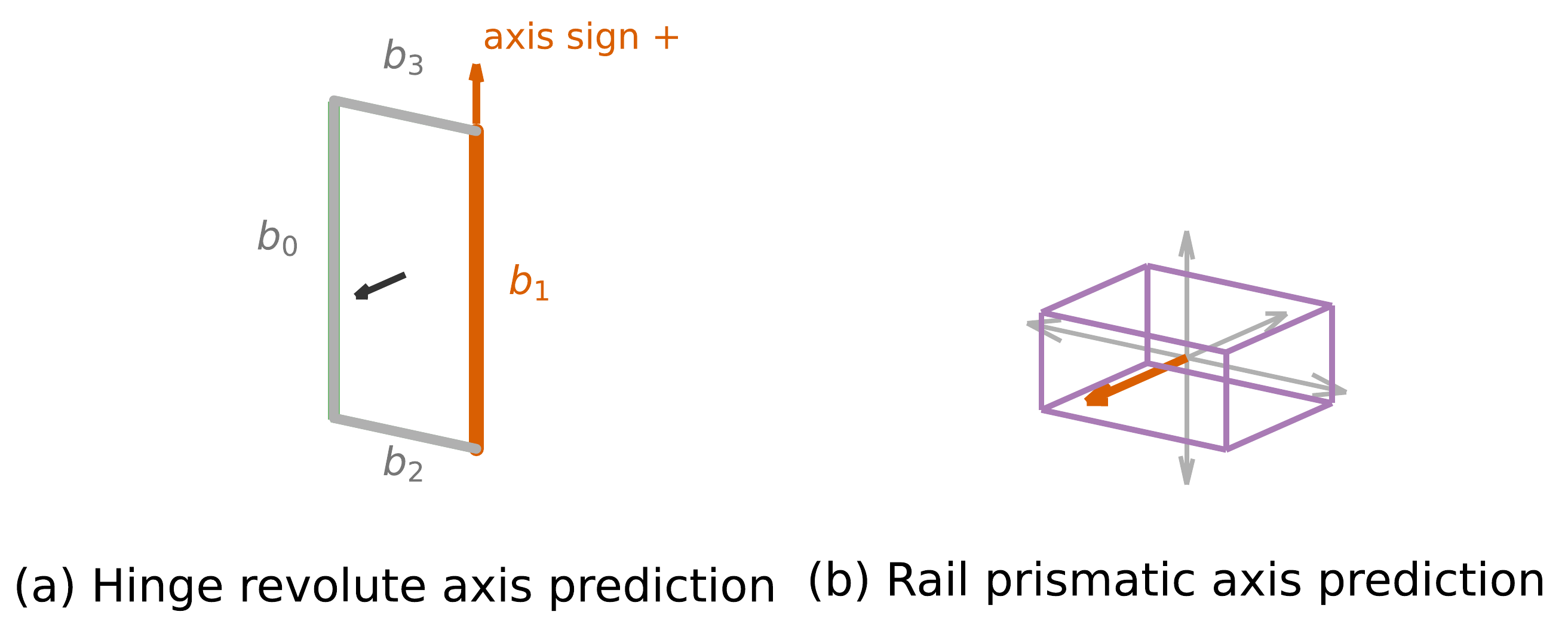}
    \caption{Motion prediction on hinge and rail functional edges. For hinge motion prediction, the revolute axis is classified as one of the four bounding box borders (gray borders) at the thin edge (black arrow), with additional sign prediction for the revolving direction. For rail edge prismatic axis prediction, the axis is classified between $\{\pm x, \pm y, \pm z\}$.}
    \label{fig:supp_motion_params}
\end{figure}
Per slot, the decoder predicts node existence, category, centroid, and box half-extents. Per slot pair, it classifies the connecting edge into the five types from the two slot features and their pairwise geometric terms. For motion, GraFu classifies rather than
regresses: a hinge's rotation axis is a 4-way border choice plus a 2-way axis sign in the door's thin-axis-aligned local frame, a drawer's prismatic axis is a 6-way signed axis (\Cref{fig:supp_motion_params}). The discrete prediction serves only as a coarse initialization and is subsequently grounded by the geometry realization stage.

After all nodes are inferred, Hungarian matching is performed between predicted slots and ground-truth nodes, allowing loss calculation. The loss formulation and data augmentation are detailed in their respective sections of this supplement.
\end{revblock}

\section{Loss Function}

Given a training pair $(G_u, G_f)$ where $G_u$ is the non-functional input graph and $G_f$ is the ground-truth functional graph, GraFu decodes a slot graph $\hat{G}=\{\hat{s}_i\}_{i=1}^{M}$ with $M$ output slots. The first $N$ slots
are \emph{anchored} to the input nodes and the next $K$ are \emph{free} queries; the remaining $M-N-K$ slots are zero-padded (and supervised to be null). We optimize a weighted sum of per-slot, per-pair, and motion losses, where the per-pair terms are evaluated after Hungarian matching between the predicted slot set and the GT node set.

\subsection{Per-slot Loss}

\paragraph{Set assignment.}
All $M$ slots are matched jointly to the GT nodes by the Hungarian
assignment 
\begin{equation}
    \sigma = \arg\min_{\sigma}\sum_{i}\big[\,
  \lambda_{\mathrm{cls}}\,\mathrm{CE}(\hat{m}_i, m_{\sigma(i)})
  + \lambda_{\mathrm{pos}}\,\|\hat{c}_i - c_{\sigma(i)}\|_2\,\big]
\end{equation}
with $(\lambda_{\mathrm{cls}}, \lambda_{\mathrm{pos}}) = (1, 2)$, where $m_i, c_i$ denote the GT category and centroid. The correspondence between the input nodes and the first $N$ slots is preserved by the anchor supervision below, which makes the identity match cost-optimal and leaves the GT nodes absent from the input to be claimed by the free slots.

\paragraph{Attribute supervision.}
The attributes of each matched slot, including its category, centroid, and bounding box, should conform to the ground truth. Correct category prediction is essential for functional-graph reasoning, while centroid and box alignment keeps the Hungarian assignment faithful and stabilizes training. Three losses enforce this over the matched pairs:
\begin{align}
  \mathcal{L}_{\mathrm{mat}}  &= \mathrm{CE}\big(\hat{m}_i,\, m_{\sigma(i)}\big), \\
  \mathcal{L}_{\mathrm{cent}} &= \big\|\hat{c}_i - c_{\sigma(i)}\big\|_2^2, \\
  \mathcal{L}_{\mathrm{bbox}} &= \big\|\hat{h}_i - h_{\sigma(i)}\big\|_2^2,
\end{align}
where $\hat{h}_i, h_i$ are the predicted and GT box half-extents, and each term is averaged over the matched slots of a batch.

The attribute loss combines the three terms as
\begin{equation}
    \mathcal{L}_{\mathrm{attr}} = \mathcal{L}_{\mathrm{mat}}
+ \mathcal{L}_{\mathrm{cent}} + 0.5\,\mathcal{L}_{\mathrm{bbox}}
\end{equation}

\paragraph{Existence supervision.}
Existence supervision is what lets GraFu fire new nodes and hence perform structural completion. Three terms address it at different granularities. A general existence loss applies to all $M$ slots,
\begin{equation}
  \mathcal{L}_{\mathrm{exist}} =
  \mathrm{BCE}\big(\hat{e}_i,\, \mathbbm{1}[i \text{ anchored or matched}]\big),
\end{equation}
Because anchored slots and padding slots occupies a large portion in all node sets, a dedicated free-slot term is added to signify existence signals from the $K$ free slots. $$\mathcal{L}_{\mathrm{free}} =
\mathrm{BCE}_{\mathrm{pw}}(\hat{e}_i, \cdot)\,,\ i \notin \mathrm{anchored}$$ with a positive-class weight ($pw$) of at least $10$; this is the model's primary signal that missing nodes must be added back. Finally, a cardinality regularizer calibrates the \emph{total} amount of completion,
\begin{equation}
  \mathcal{L}_{\mathrm{count}} =
  \Big(\big|\textstyle\sum_{i \in \mathrm{free}} \sigma(\hat{e}_i)
  - N^{\mathrm{gt}}_{\mathrm{free}}\big|\Big)^{1/2},
\end{equation}
where $N^{\mathrm{gt}}_{\mathrm{free}}$ counts the GT nodes not covered by any anchored slot. It penalizes over-firing and under-firing induced by incorrect free slots prediction.

The existence loss combines the three granularities as
\begin{equation}
    \mathcal{L}_{\mathrm{ext}} = \mathcal{L}_{\mathrm{exist}}
+ 3\,\mathcal{L}_{\mathrm{free}} + \mathcal{L}_{\mathrm{count}}
\end{equation}

\paragraph{Anchor supervision.}
Input nodes must be fully preserved through the graph-to-graph
translation. Rather than hard-constraining the assignment, we let GraFu learn this softly through two pose-consistency terms that pull each anchored slot back onto its \emph{input-node} pose,
\begin{equation}
\begin{aligned}
  \mathcal{L}_{\mathrm{anchor}} &=
  \big\|\hat{c}_i - c^{\mathrm{in}}_i\big\|_2^2, \\
  \mathcal{L}_{\mathrm{anchor\text{-}bbox}} &=
  \big\|\hat{h}_i - h^{\mathrm{in}}_i\big\|_2^2,
\end{aligned}
\qquad i \in \mathrm{anchored},
\end{equation}
whose targets are the input poses themselves, independent of the
assignment. 

The anchor loss is
\begin{equation}
    \mathcal{L}_{\mathrm{anc}} = 0.3\,(\mathcal{L}_{\mathrm{anchor}}
+ \mathcal{L}_{\mathrm{anchor\text{-}bbox}})
\end{equation}
and the per-slot loss aggregates the three groups,
\begin{equation}
\mathcal{L}_{\mathrm{slot}} = \mathcal{L}_{\mathrm{attr}}
+ \mathcal{L}_{\mathrm{ext}} + \mathcal{L}_{\mathrm{anc}}
\end{equation}

\subsection{Per-pair Loss}

\paragraph{Edge classification.}
For every pair $(i,j)$ of real slots, the decoder emits an edge-type distribution $\hat{r}_{ij}$ over five classes \{\texttt{contact}, \texttt{hinge}, \texttt{rail}, \texttt{attached}, \texttt{null}\}, with the GT edge matrix constructed through the matched correspondence $\sigma$. Because the fully connected pair set is dominated by \texttt{null} pairs while the functional edges are rare, the edge loss is a class-weighted cross-entropy,
\begin{equation}
  \mathcal{L}_{\mathrm{edge}} =
  \mathrm{CE}_{\omega}\bigl(\hat{r}_{ij},\; r_{ij}\bigr),
  \qquad \omega = (1,\, 5,\, 5,\, 3,\, 0.2),
\end{equation}
which suppresses the trivial null majority and emphasizes the \texttt{hinge}, \texttt{rail}, and \texttt{attached} classes that carry the functionalization.

\paragraph{Exclusive-relation supervision.}
Two relations are exactly one by nature: a handle attaches to exactly one door or drawer, and a door hinges onto exactly one static panel. The per-pair cross-entropy above calibrates each pair in isolation but never compares candidates against one another, whereas inference must select a single winner by an argmax over candidates. Two dedicated heads therefore score each candidate pair, $\rho_{ij}$, alongside a learned abstain logit $\rho_{\varnothing}$, and are supervised with a cross-entropy over the across-candidate softmax that inference selects with:
\begin{equation}
\begin{aligned}
  \mathcal{L}_{\mathrm{parent}} &=
  -\log \frac{\exp(\rho_{i j^{*}})}
  {\sum_{j \in \mathcal{C}_i} \exp(\rho_{ij}) + \exp(\rho_{\varnothing})},
\end{aligned}
\end{equation}
where $\mathcal{C}_i$ are the candidate parents (door and drawer slots)
of handle slot $i$ and $j^{*}$ its GT parent; $\mathcal{L}_{\mathrm{ht}}$
is the analogue on the door side, over candidate static panels. Beyond
ranking, the abstain logit gives the selection a calibrated ``no
parent'' option for handles whose parent is absent from the prediction.
The per-pair loss combines the terms as
$\mathcal{L}_{\mathrm{pair}} = \mathcal{L}_{\mathrm{edge}}
+ 0.5\,\mathcal{L}_{\mathrm{parent}} + 0.5\,\mathcal{L}_{\mathrm{ht}}$.

\subsection{Motion Loss}
For each GT \texttt{hinge} edge and \texttt{rail} edge, the motion heads predict pose-invariant joint attributes as discrete classifications: the 4-way hinge border $\hat{f}^{(4)}_{ij}$, the 2-way axis sign
$\hat{\epsilon}^{(2)}_{ij}$, and the
6-way signed rail axis $\hat{a}^{(6)}_{ij}$ for drawers:
\begin{equation}
  \mathcal{L}_{\mathrm{motion}} =
  \mathrm{CE}\bigl(\hat{f}^{(4)}_{ij}, f_{ij}\bigr)
  + \mathrm{CE}\bigl(\hat{\epsilon}^{(2)}_{ij}, \epsilon_{ij}\bigr)
  + \mathrm{CE}\bigl(\hat{a}^{(6)}_{ij}, a_{ij}\bigr),
\end{equation}
where the first two terms apply to hinge edges and the third to rail edges, each averaged over its edges in a batch.

\section{Loss Ablation}
\label{sec:loss_ablation}

\paragraph{Setup.}
We probe the contribution of each supervisory signal by holding out one
loss component (or a tightly coupled pair, when two losses jointly
supervise the same quantity) on top of the winning configuration and
retraining from scratch under identical optimisation settings
($85$ epochs, cosine-annealed learning rate with $T_{\max}{=}450$, same
train/val/test split as the main model). The six variants we evaluate
are:
\begin{itemize}
\item $-w_{\mathrm{ht}}$: removes $\mathcal{L}_{\mathrm{hinge\_target}}$.
This loss trains the directed head that, for each door, picks exactly
one static mounting panel; removing it disables the door-to-panel
selection signal.
\item $-w_{\mathrm{parent}}$: removes $\mathcal{L}_{\mathrm{parent}}$.
This loss trains the directed head that, for each handle, picks exactly one parent door or drawer; removing it disables the handle-to-parent selection signal.
\item $-w_{\mathrm{cent}}$: removes the centroid loss inside
$\mathcal{L}_{\mathrm{attr}}$. This loss supervises slot centroids; removing it leaves bbox half-extents as the only direct geometry supervision.

\item $-w_{\mathrm{bbox}}{+}w_{\mathrm{anchor\_bbox}}$: removes both bbox losses (free-slot and anchor-side). These jointly carry all bbox supervision; removing them leaves the model with no signal about part size.

\item $-w_{\mathrm{anchor}}{+}w_{\mathrm{anchor\_bbox}}$: removes both anchor pose losses (centroid and bbox). These jointly tie anchored slots to their input-node poses; removing them lets the encoder freely drift the identity transform on existing parts.

\item $-w_{\mathrm{free\_exist}}{+}w_{\mathrm{count}}$: removes the two free-slot firing supervision. Together, they regulate when and how many free slots fire; removing them frees the model to predict any number of hallucinated nodes.
\end{itemize}

We pair each loss with another only when both supervise the same
underlying quantity from different angles, so that the ablation isolates
\emph{a function} (e.g.\ bbox supervision, anchor-pose preservation, or
hallucination control) rather than an arbitrary single weight.
Table~\ref{tab:loss_ablation} reports the resulting test-set metrics against the full setup

\paragraph{Conclusion.}
The full setup is the best overall configuration. Looking across
Table~\ref{tab:loss_ablation}, it attains the highest hinge edge F1, hinge axis-sign accuracy, and joint motion accuracy (the signature metrics for our task), and ranks at-or-near best on the remaining metrics. Importantly, no single
ablation improves more than one metric in isolation while leaving the others unaffected, ruling out the possibility that any loss is redundant.

\section{Data Augmentation}
\label{sec:augmentation}

\begin{figure}
    \centering
    \includegraphics[width=\linewidth]{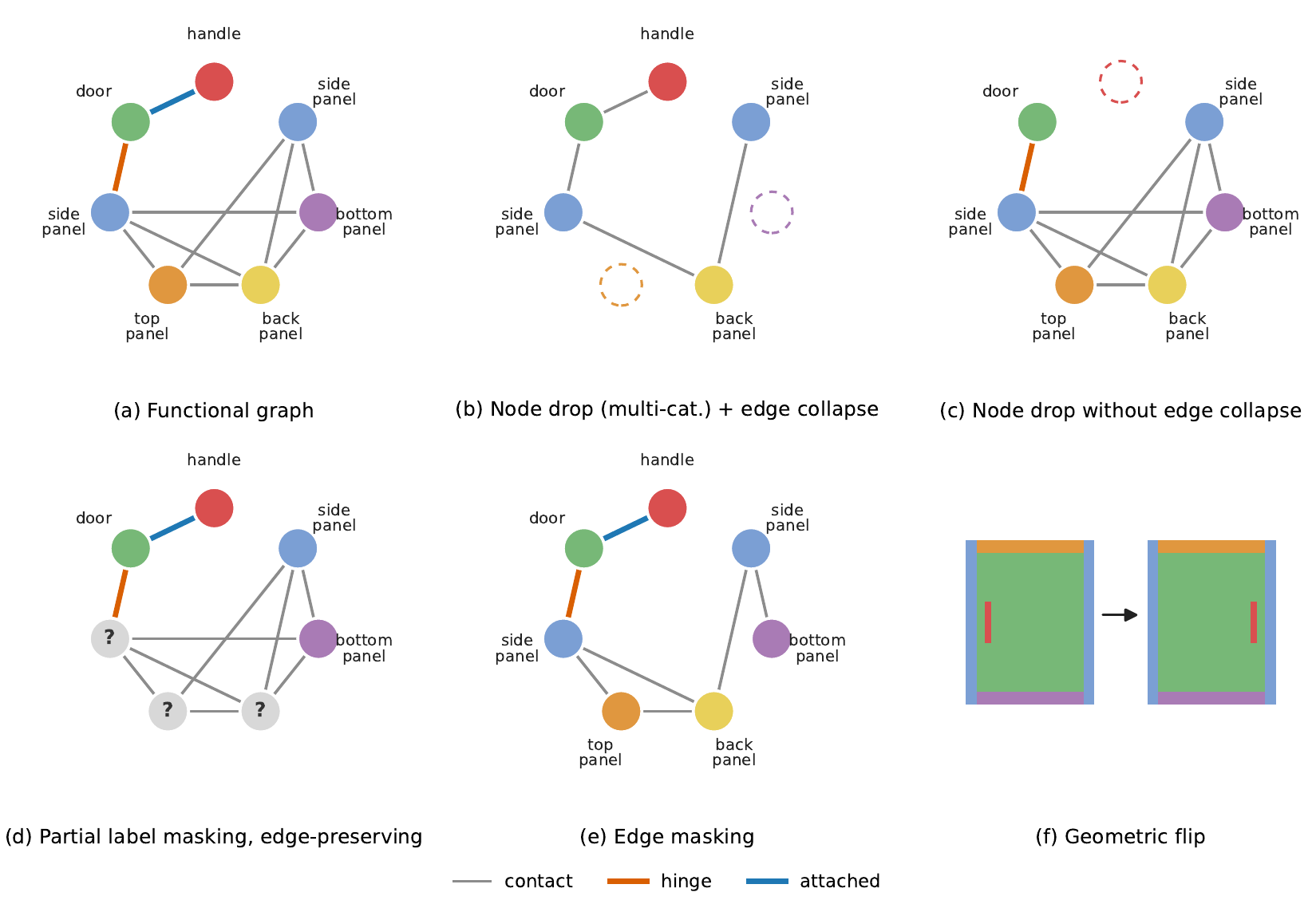}
    \caption{\rev{Example data augmentation strategies.}}
    \label{fig:data aug}
\end{figure}

Beyond the natural $(G_u, G_f)$ pairs obtained from FurFun-233 (where $G_u$ is the rest-state contact-only graph and $G_f$ the functional counterpart), we synthesize additional input variants by corrupting the functional graph $G_f$ via various structure-corrupting strategies plus a random geometric flip. Illustrative examples are shown in \Cref{fig:data aug}. Per training step, a strategy is sampled from a weighted categorical distribution (Table~\ref{tab:aug}); the sampled corruption is then applied to $G_f$ to produce a synthetic $G_u^{\mathrm{aug}}$, while the supervision target remains the original $G_f$. Empirically, we augment the unfunctional graphs to 17 folds for training.

\paragraph{Category drops.}
Five strategies remove all nodes of a specific structural category (handles, top panels, bottom panels, doors, drawers). Each forces the free-slot existence and material heads to recover the missing category from the surrounding contact graph and the global pooled summary fed into the free queries. A sixth strategy removes two or three of the above categories simultaneously, training the decoder to handle compound completion where multiple structural roles are missing at once. A seventh strategy randomly removes a handful of arbitrary nodes regardless of category. In all cases the dropped node's incident edges are deleted with it, and the surviving edges are relabeled as plain \texttt{contact}.

\paragraph{Label-masking corruptions.}
Two strategies anonymize the input labels at complementary severities. The first replaces \emph{every} node's category with the \emph{unknown} token, and with all of its edges relabelled as plain \texttt{contact}. The second masks a random 30--60\% of the categories while preserving the surviving edges' functional types, pairing partial semantics with an intact relational structure. Both train the material head to recover part categories from geometry and topology without explicit semantic labels.

\paragraph{Edge-preserving partial-functional drops.}
Three strategies contribute to \emph{edge-preserving} family: every surviving edge keeps its functional type, training the model on partial-functional inputs. They are:

\begin{itemize}\setlength\itemsep{1pt}

\item \texttt{mask\_random\_edges\_keep\_types} removes a random 25--50\% of the edges while keeping all nodes;

\item \texttt{drop\_nodes\_keep\_edge\_types} removes a random 30--50\% of the nodes, keep the remaining edges;

\item \texttt{fully\_functional\_drop} removes a random number of non-dynamic nodes and their incidental edges such that none of the functional edges are removed. All remaining edges are preserved. 
\end{itemize}

\paragraph{Geometric flip.}
Independent of the corruption strategy, with probability $p=0.5$ we apply a random axis-aligned mirror across $X$ or $Y$. Coordinates and bbox extents are negated on the mirrored axis, point-cloud samples are flipped, and the motion labels are remapped. The mirror doubles the effective data and enhances the geometric diversity without collecting new data.

\paragraph{Sampling weights.}
Table~\ref{tab:aug} lists the categorical sampling weight of each strategy; weights are renormalized to a probability distribution at sampling time.

\begin{table*}[t]
\centering
\small
\caption{Loss hold-out ablation on the FurFun test split. Each column zeros one loss term (or pair). \textbf{Bold} = best per row; $\downarrow$ / $\uparrow$ indicates desirable direction. Every variant degrades at least one signature metric, and no single loss is dispensable.}
\label{tab:loss_ablation}
\begin{tabular}{l c c c c c c c}
\toprule
\textbf{Metric}
& \textbf{Full}
& $-w_{\mathrm{ht}}$
& $-w_{\mathrm{parent}}$
& $-w_{\mathrm{cent}}$
& $-w_{\mathrm{bbox+anc.bbox}}$
& $-w_{\mathrm{anc+anc.bbox}}$
& $-w_{\mathrm{free\_exist+count}}$ \\
\midrule
Node existence F1   $\uparrow$        &0.972 & 0.959 & \textbf{0.974} & 0.961 & 0.970 & 0.969 & 0.964 \\
Material accuracy   $\uparrow$        & 0.957 & 0.947 & 0.958 & 0.949 & \textbf{0.961} & 0.954 & \textbf{0.961} \\
Centroid MAE (m)    $\downarrow$      & 0.060 & 0.048 & \textbf{0.046} & 0.092 & 0.060 & 0.079 & 0.056 \\
Bbox MAE (m)        $\downarrow$      & 0.081 & 0.078 & \textbf{0.069} & 0.114 & \emph{0.298} & 0.108 & 0.075 \\
Edge acc. (real)    $\uparrow$        & 0.770 & 0.637 & 0.649 & 0.646 & 0.725 & 0.704 & \textbf{0.777} \\
Edge F1 contact     $\uparrow$        & 0.884 & 0.836 & 0.858 & 0.835 & \textbf{0.896} & 0.844 & 0.892 \\
Edge F1 hinge       $\uparrow$        & \textbf{0.656} & \emph{0.409} & 0.567 & 0.565 & 0.573 & 0.580 & 0.652 \\
Edge F1 rail        $\uparrow$        & 0.447 & 0.377 & 0.407 & 0.395 & 0.431 & 0.398 & \textbf{0.479} \\
Edge F1 attached    $\uparrow$        & 0.855 & 0.778 & \emph{0.004} & 0.749 & 0.791 & 0.835 & 0.823 \\
Hinge border-4 acc. $\uparrow$        & 0.922 & 0.918 & 0.930 & 0.877 & 0.926 & 0.866 & \textbf{0.940} \\
Hinge axis-sign acc.$\uparrow$        & \textbf{0.975} & 0.925 & 0.899 & 0.897 & 0.940 & 0.925 & 0.954 \\
Rail axis acc.      $\uparrow$        & 0.997 & 0.992 & 0.992 & \textbf{1.000} & \textbf{1.000} & \textbf{1.000} & 0.996 \\
\textbf{Joint motion} (hb4$\times$has) $\uparrow$
                                      & \textbf{0.899} & 0.849 & 0.836 & 0.786 & 0.870 & 0.801 & 0.897 \\
\bottomrule
\end{tabular}
\end{table*}

\begin{table}[t]
\centering
\small
\caption{Categorical sampling weights for augmentation strategies.}
\label{tab:aug}
\begin{tabular}{lll}
\toprule
Family & Strategy & Weight \\
\midrule
Category drop   & drop\_handles                    & 0.15 \\
                & drop\_top                        & 0.15 \\
                & drop\_bottom                     & 0.10 \\
                & drop\_doors                      & 0.08 \\
                & drop\_drawers                    & 0.08 \\
                & drop\_multi\_cat                 & 0.10 \\
                & random                           & 0.05 \\
\midrule
Label masking   & fully\_anonymized                & 0.05 \\
     & mask\_materials\_keep\_edges     & 0.04 \\
\midrule
Edge-preserving & mask\_random\_edges\_keep\_types & 0.10 \\
                & drop\_nodes\_keep\_edge\_types   & 0.05 \\
                & fully\_functional\_drop          & 0.20 \\
\bottomrule
\end{tabular}
\end{table}
\begin{revblock}
\paragraph{Validation} The 17-fold data augmentation is empirically chosen based on FurFun validation set performance. Here we conduct a post-validation on PN-M dataset, benchmarking on its GT-aligned motion accuracy. We train GraFu with 0, 5, 10 and 20-fold data augmentation and test on PN-M testset. The motion accuracy are $55.78\%, 84.08\%, 85.8\%$ and $85.59\%$, respectively. The accuracy change is visualized in \Cref{fig:fold-ablation}. It can be seen that data augmentation provides substantial support in GraFu learning from 5-fold, and plateaued from 10 fold. Our choice of 17-fold is inbetween this plateaued range. 

\begin{figure}
    \centering
    \includegraphics[width=0.8\linewidth]{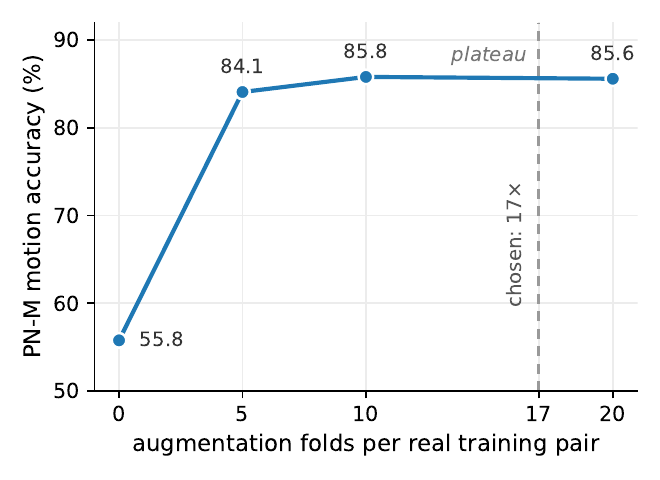}
    \caption{Data augmentation fold ablation experiment.}
    \label{fig:fold-ablation}
    \vspace{-1 em}
\end{figure}
\end{revblock}

\section{Geometric Fix Algorithms}
\label{sec:geometric_fix}
Given the functional graph predicted by GraFu, the geometric fixing stage grounds it in 3D. \emph{Dynamic} functionalization inserts real mechanism templates for every predicted \texttt{hinge} and \texttt{rail} edge, rectifying the motion axis in the process; \emph{static} functionalization repairs non-moving defects, including missing countertops, absent handles, and empty interiors.

\subsection{Dynamic Functionalization}
Rather than inferring mechanical parameters from data, we use animated hinge and rail templates, each annotated once under a shared protocol. This turns the continuous placement problem into a \emph{discrete selection} of a template per installation condition, followed by \emph{geometric fitting}.

\subsubsection{Template Annotation}
\begin{figure}
    \centering
    \includegraphics[width=\columnwidth]{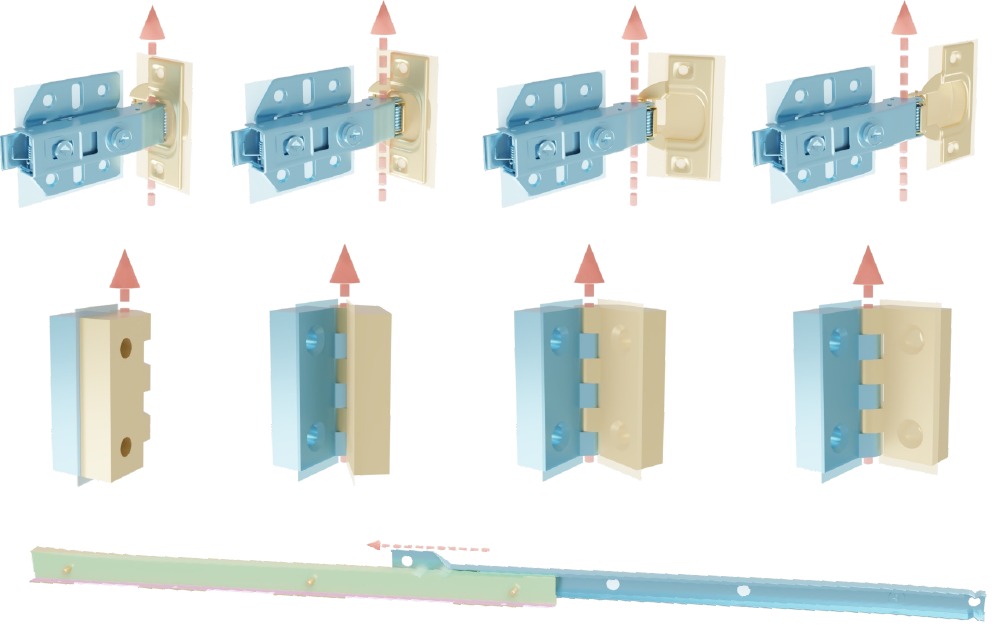}
    \caption{Mechanical part annotations. Row 1: P-hinge with non-linear motion; Row 2: C-hinge with linear revolute motion. Cyan and gold indicate geometry and snap plane for static and dynamic parts, respectively. Row 3: Sliding rail. Magenta and lime indicate two snapping planes.}
    \label{fig:annotation}
\end{figure}

All connectors follow the same annotation protocol, so the annotations transfer to new template instances. Each template is partitioned into a \emph{static end}, which contacts the static mounting panel, and a \emph{dynamic end}, which mounts on the movable part. The annotation comprises two elements (\Cref{fig:annotation}): a \textbf{motion axis}
and a pair of \textbf{snap planes} $\{\Pi_s, \Pi_d\}$, one per end.

For hinges, the motion axis lies on the pin, oriented by the right-hand rule so that the positive direction indicates counter-clockwise rotation; for rails, it indicates the sliding direction. Each snap plane is a zero-thickness rectangle covering its mounting region; during snapping, its distance to the target mounting surface serves as the fitting objective. 

These elements constitute the minimum annotation; templates with specific installation rules carry additional planes, such as the magenta plane in \Cref{fig:annotation}, which encodes that rail installation requires exact bottom support beneath the drawer.

\subsubsection{Revolute Motion: Hinge Attachment}
\begingroup
    \setlength{\columnsep}{1em}
    \setlength{\intextsep}{0em}
    
Physically plausible revolute motion requires bilateral contact: the hinge must meet both the static part $\mathcal{S}$ and the movable part $\mathcal{M}$ along geometrically compatible surfaces.

We first select a compatible hinge class from the spatial relation between $\mathcal{M}$ and $\mathcal{S}$, highlighted in Figure~3 of the main paper. Each
\begin{wrapfigure}[6]{r}{0.5\linewidth}
          \centering
          \includegraphics[width=\linewidth]{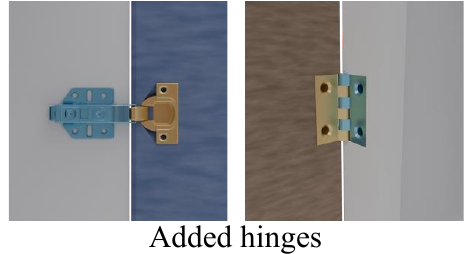}
        \end{wrapfigure} 
class implies a different pair of snapping targets: a C-hinge seeks contacting faces, an F-hinge seeks flush faces, and a P-hinge seeks perpendicular faces, all detected on bounding-box proxies of $\mathcal{M}$ and $\mathcal{S}$. When several classes apply, we prefer C- over F- over P-hinges.

Insertion is initialized from a motion axis
$\mathbf{a}=(\mathbf{o},\mathbf{d})$, 
with origin $\mathbf{o}$ and direction $\mathbf{d}$. This can be read from GraFu predictions, GT annotations, or other motion-prediction networks
(e.g.,~\cite{li2025particulatefeedforward3dobject,liu2024singapo}).
The axis serves only as initialization and a coarse estimate suffices.
\endgroup

\begin{figure}
    \centering
    \includegraphics[width=\linewidth]{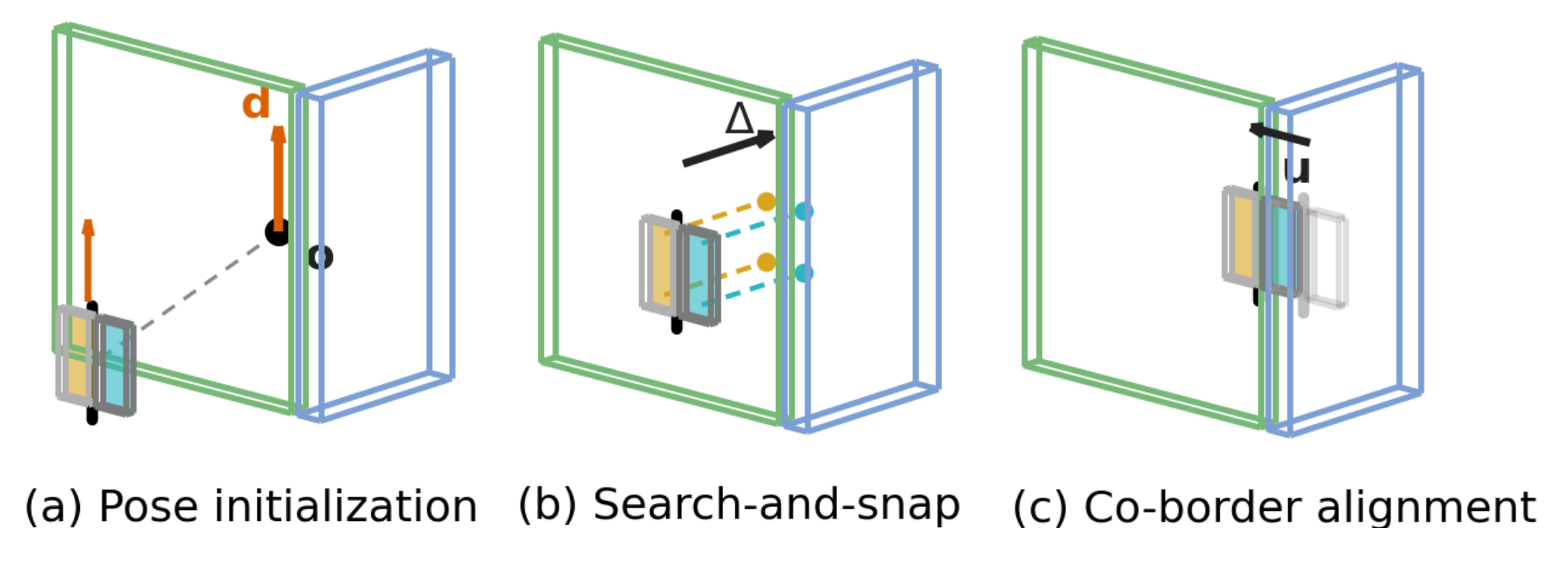}
    \caption{\rev{Hinge insertion procedures}}
    \label{fig:hinge_install}
\end{figure}

Fitting proceeds in three steps, shown in \Cref{fig:hinge_install}.

\paragraph{Pose Initialization.} The template is translated so its pin passes through $\mathbf{o}$ and rotated by the minimal rotation aligning its motion axis with $\mathbf{d}$. 
A flip about the axis is applied when needed so that the $\Pi_s$ faces $\mathcal{S}$ and the $\Pi_d$ faces $\mathcal{M}$, and the template is scaled so that the hinge leaves do not protrude from their mounting faces.

\paragraph{Search and Snap.} We then snap the annotated planes onto the actual meshes. Let $\mathbf{x}_s, \mathbf{x}_d$ denote the snap-plane centroids, $\mathbf{n}_s, \mathbf{n}_d$ their normals, and $\mathbf{p}_s, \mathbf{p}_m$ the contact points located on $\mathcal{S}$ and $\mathcal{M}$ by casting rays along the plane normals. The snap translation $\boldsymbol{\delta}$ solves
\begin{equation}
  \boldsymbol{\delta}\!\cdot\!\mathbf{n}_s=(\mathbf{p}_s-\mathbf{x}_s)\!\cdot\!\mathbf{n}_s,\quad
  \boldsymbol{\delta}\!\cdot\!\mathbf{n}_d=(\mathbf{p}_m-\mathbf{x}_d)\!\cdot\!\mathbf{n}_d,\quad
  \boldsymbol{\delta}\!\cdot\!\mathbf{d}=0.
  \label{eq:hinge_refine}
\end{equation}
The two contact constraints leave $\boldsymbol{\delta}$ free along $\mathbf{d}$, which indicates the hinge's position along the door border. The third constraint removes this freedom; the axial position is instead set explicitly, placing the $i$-th of $N$ hinges at height $i/(N{+}1)\cdot h$ on a door of height $h$.

\paragraph{Co-bordering.} When $\mathbf{n}_s \parallel \mathbf{n}_d$, the two contact constraints in \cref{eq:hinge_refine} become linearly dependent, leaving one more free direction $\mathbf{u}=\mathbf{n}_s\times\mathbf{d}$,
in the mounting plane and perpendicular to the axis. We resolve it by aligning borders. Writing $e_{\mathbf{o}}(X)$ for the extreme of
$\{\,v\cdot\mathbf{u} : v\in X\,\}$ nearest the pin projection
$\mathbf{o}\cdot\mathbf{u}$, the correction is
\begin{equation}
  \Delta_{\text{co-border}}
    = \bigl(e_{\mathbf{o}}(\partial\mathcal{M}) - e_{\mathbf{o}}(\partial\Pi_d)\bigr)\,\mathbf{u},
  \label{eq:coborder}
\end{equation}
which brings the dynamic snap plane's pin-side edge onto the movable part's hinge-side border, doing so places the pin exactly at the shared edge of the movable and static parts.

\subsubsection{Prismatic Motion: Sliding Rail Attachment}
\begin{figure}
    \centering
    \includegraphics[width=\linewidth]{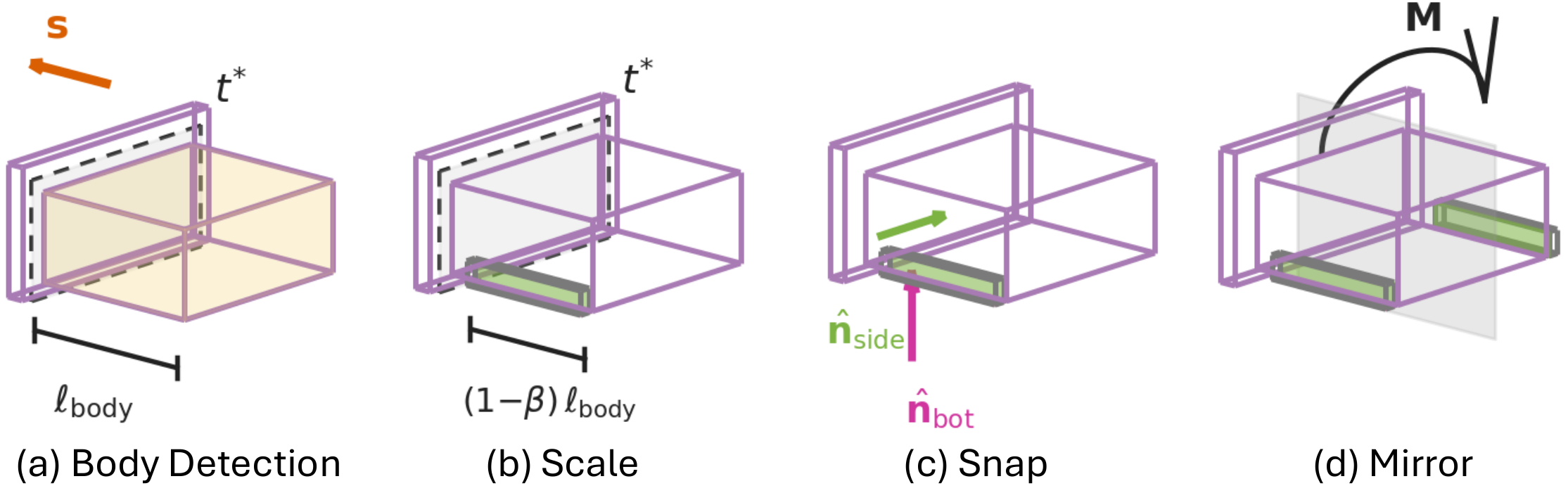}
    \caption{\rev{Rail insertion procedures}}
    \label{fig:rail_install}
\end{figure}

A drawer slides on a pair of rails mounted between its container body and the cabinet walls. Placing them reduces to three questions: how long the rail may be? Where does it contact the drawer? How does the pair stay symmetric? The proposed fitting solution proceeds in four steps (\Cref{fig:rail_install}), followed by an optional support block.


\begingroup
    \setlength{\columnsep}{1em}
    \setlength{\intextsep}{0em}
\paragraph{Body-Region Detection.} The rail length should be proportional to the drawer length, capped behind the pulling board. The rail length 
 \begin{wrapfigure}[6]{r}{0.25\linewidth}
          \centering
          \includegraphics[width=\linewidth]{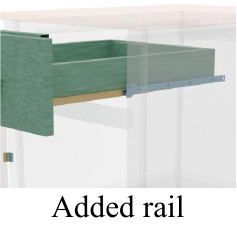}
        \end{wrapfigure}
determines the motion range of the drawer, and it should conform to the container's length. The first step is to determine the container body region. A dense ray grid over the drawer's flank yields its outer-width profile $W(t)$ along the sliding axis $\mathbf{s}$
(\Cref{fig:rail_install}a). The body width is the profile's median, and the pulling board contributes only as outliers.
\begin{equation}
  T_{\text{body}}
    = \bigl\{\,t : |W(t)-\operatorname{med}_{t'}W(t')| < \tau_W \,\bigr\},
  \label{eq:body_region}
\end{equation}
with $\tau_W$ a geometric tolerance. Its boundary $t^{*}$ toward the pulling board separates body from board, and $\ell_{\text{body}}$ is the body's length along $\mathbf{s}$.

\endgroup
\paragraph{Scaling.} The template is rotated minimally to align its prismatic axis with $\mathbf{s}$, rolled about $\mathbf{s}$ so its static snap plane faces the cabinet wall, and scaled uniformly to $(1-\beta)\,\ell_{\text{body}}$ (\Cref{fig:hinge_install}b). Empirically, we set $\beta=0.1$. The end of the rail template is aligned with the drawer end.

\paragraph{Snapping.} Exact contact is recovered by
directional from snap plane normals, analogy to hinge template snapping (\Cref{fig:hinge_install}c). For rails with bottom support, the equation write as
\begin{equation}
  \mathbf{n}_{\text{bot}}=+\hat{\mathbf{z}},\qquad
  \mathbf{n}_{\text{side}}=\pm(\mathbf{s}\times\hat{\mathbf{z}}),
  \label{eq:rail_priors}
\end{equation}
The dynamic snap plane contacts the drawer's flanks and the bottom-support plane contacts the underside of the drawer bottom. Every cast restricted to $T_{\text{body}}$ so pulling-board geometry cannot bias the fit. Center-mount templates carry no bottom support plane and are centered vertically on the drawer side instead.

\paragraph{Symmetric Mirroring.} We optimize one rail only; the other is its reflection across the drawer's sagittal plane
(\Cref{fig:rail_install}d),
\begin{equation}
  \mathbf{M}_{\text{mirror}}
    = \mathbf{I}-2\,\mathbf{n}_{\perp}\mathbf{n}_{\perp}^{\top},
  \qquad \mathbf{n}_{\perp}=\mathbf{s}\times\hat{\mathbf{z}},
  \label{eq:rail_mirror}
\end{equation}
which enforces exact bilateral symmetry.

\paragraph{Lateral Support Block.} A rail needs a mounting point behind its static snapping plane, which the cabinet does not always provide flush. When a lateral gap remains, a thin support block is extruded from the static snap plane to the wall; rails already contacting with the wall skip this step.

\subsection{Static Functionalization}
Static repairs missing structural parts such as countertops, handles, and interiors such as shelves and dividers. 

\subsubsection{Adding Countertops}

\begingroup
\setlength{\columnsep}{1em}
\setlength{\intextsep}{0em}
    We frequently observe base cabinets in PartNet-Mobility~\cite{xiang2020sapien} that are modeled without an explicit countertop. This is often reasonable when the cabinet is intended to be placed beneath a shared counter surface or paired with appliances (\textit{e.g.}, a stove). However, when the asset is considered in isolation, the missing top surface makes it unusable as a standalone cabinet. We therefore treat \emph{missing countertops} as a static malfunction and repair them as part of our functionalization pipeline.
    
    To check if an object has a countertop, from a top view, we look at how much space it covers on the floor and how much of that space is covered near the top of the object. 
    If a large part of the floor area is also covered near the top, we say a countertop exists. 
        \begin{wrapfigure}[7]{r}{0.55\linewidth}
          \centering
          \includegraphics[width=\linewidth]{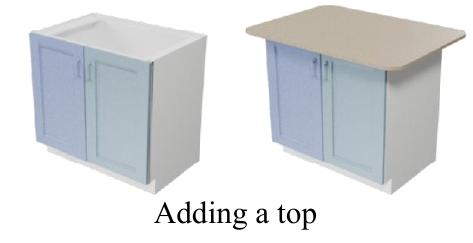}
        \end{wrapfigure}
    Algorithmically, we project the mesh onto the ground plane and create two binary occupancy maps: $\mathbf{O}_{all}$ for the entire mesh and $\mathbf{O}_{top}$ for geometry near the maximum height. 
    We then compute the coverage ratio $c = \texttt{area(}\mathbf{O}_{top}\texttt{)} / \texttt{area(}\mathbf{O}_{all}\texttt{)}$. If $c$ exceeds a threshold $\alpha$, we declare a countertop present; otherwise, it is considered missing.
    \am{When a countertop is missing, we generate one by first estimating a planar outline and then extruding it into a slab. The outline is obtained by slicing the furniture near the top, fitting an axis-aligned rectangle to the rim points (with optional overhang), and extruding it by a user-defined thickness to form the final panel.}
    
\endgroup

\subsubsection{Adding Interaction Elements}

\begingroup
    \setlength{\columnsep}{1em}
    \setlength{\intextsep}{0em}
    Interaction elements refer to geometries that enable a user to grasp and operate movable parts. We consider two complementary types. \emph{Additive} elements correspond to 
        externally attached handles that protrude from the drawer/door surface.
    \emph{Subtractive} elements correspond to \emph{recessed} grips, where a portion of the drawer/door geometry is carved to form a finger groove (with no external attachment). 

When \emph{attaching} interaction elements, our pipeline handles both additive handles and recessed grips. 
The main steps are choosing a placement region and setting handle orientation and size. 
For 
\begin{wrapfigure}[15]{r}{0.55\linewidth}
          \centering
          \includegraphics[width=\linewidth]{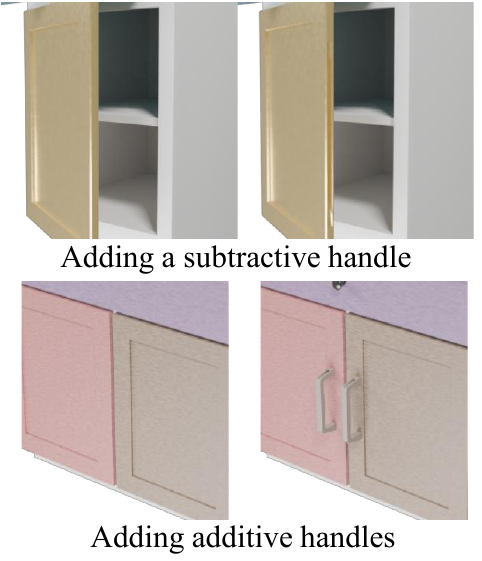}
        \end{wrapfigure}
        drawers, additive handles go at the front center, and recessed grips are carved on a chosen side. 
For doors, we use the motion axis to find the hinge side, placing handles on the opposite side, additive at mid-height with an offset, recessed carved into the panel thickness.

For additive handles, we add a small \emph{snap plane} to define the mounting position. During attachment, the handle is oriented toward the target, rays are cast from the snap plane to the surface, and the handle is moved until it sits flush. The handle is also scaled to match the part. For recessed grips, we subdivide the target face and deform vertices into a U-shaped groove, with profile (sharpness, depth, size) controllable by users.
    
\endgroup

\subsubsection{Interior Generation}
\begingroup
    \setlength{\columnsep}{1em}
    \setlength{\intextsep}{0em}
While the previous components address exterior defects, interior structure equally defines the functionality of cabinet-like furniture(e.g., wardrobes), yet is frequently missing from both human-made repositories~\cite{deitke2023objaverse,xiang2020sapien} and
automatically generated models~\cite{yang2025omnipart}. We repair it in two steps: detect the empty interior space as a set of box-shaped \emph{compartments}, then furnish selected compartments with shelves and dividers.

\paragraph{Compartment Detection.} A compartment is a maximal empty box inside the storage furniture. The predicted functional graph makes 
\begin{wrapfigure}[7]{r}{0.5\linewidth}
          \centering
          \includegraphics[width=\linewidth]{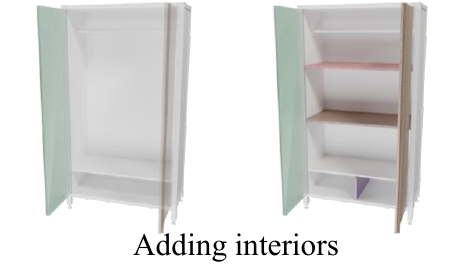}
        \end{wrapfigure}
        finding them direct: the labelled wall panels bound the interior cavity by their inner faces, and the labelled interior structure, including shelves, dividers, and drawers, marks the space already taken. Subtracting the occupied regions from the cavity and merging what remains into maximal axis-aligned boxes yields the compartments; slots too small to furnish are discarded. 

\paragraph{Panel Generation.} Each selected compartment hosts $N$ equally spaced panels of user-set thickness, horizontal for shelves and vertical for dividers, with an optional inset from the cabinet front. Generation is per compartment, so different compartments can host different configurations within the same asset.

This supports generating interiors from scratch as well as augmenting existing ones. Although we do not propose an explicit interior detector, compartment detection itself indicates whether an asset contains meaningful interior capacity. The primary limitation is the box representation: irregularly shaped interiors are not directly modeled.
\endgroup


\begin{revblock}
\section{Per-case Evaluation}
Table~2 of the main text reports quantitative results on a per-part basis. Here we complement it with a per-model evaluation on the PN-M test set.

\subsection{Motion Prediction}
\begin{figure}
    \centering
    \includegraphics[width=\linewidth]{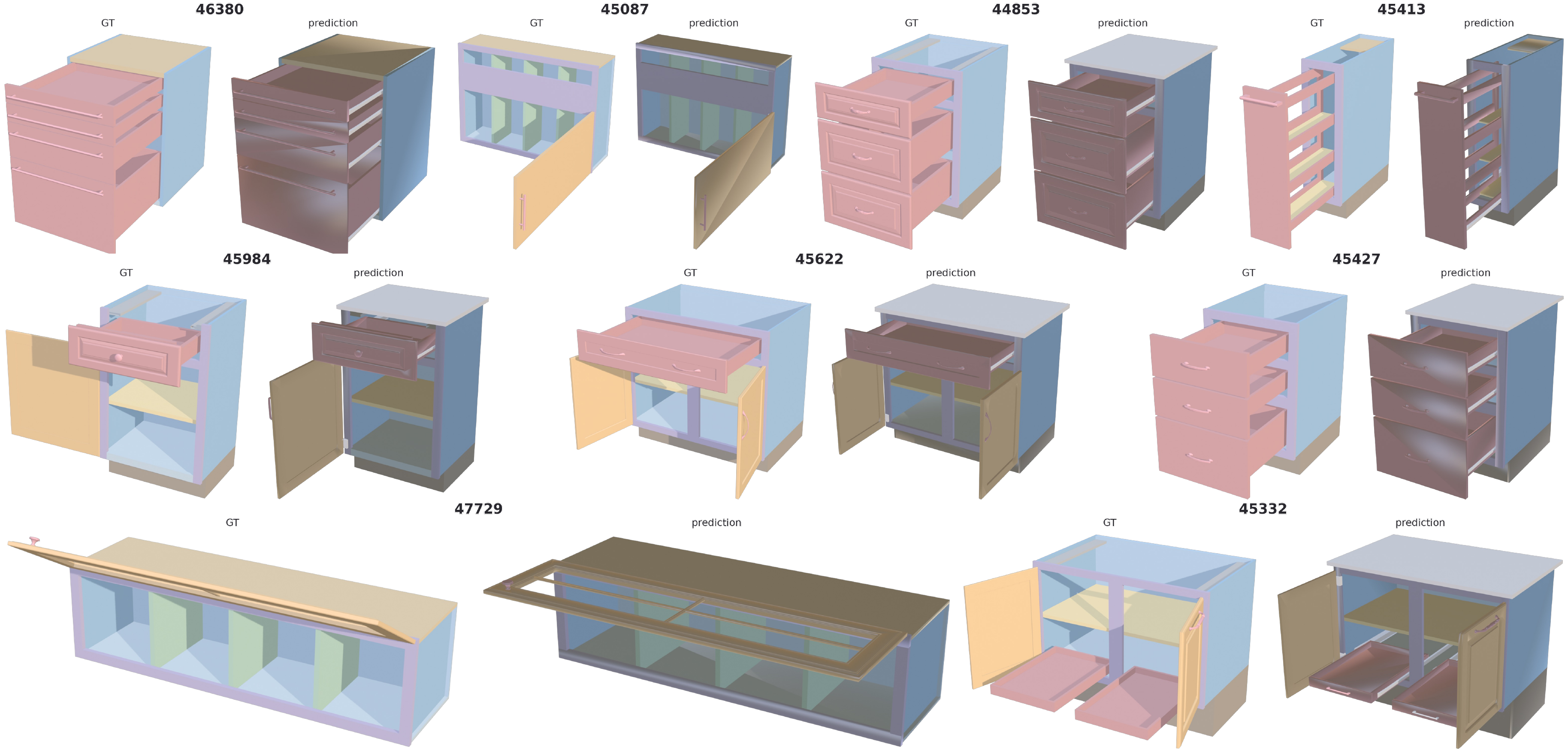}
    \caption{GT-aligned motion prediction result showcase. For each pair, GT is on the left, and the prediction is on the right. Model ID is shown above.}
    \label{fig:mp_correct}
\end{figure}

\begin{figure}
    \centering
    \includegraphics[width=\linewidth]{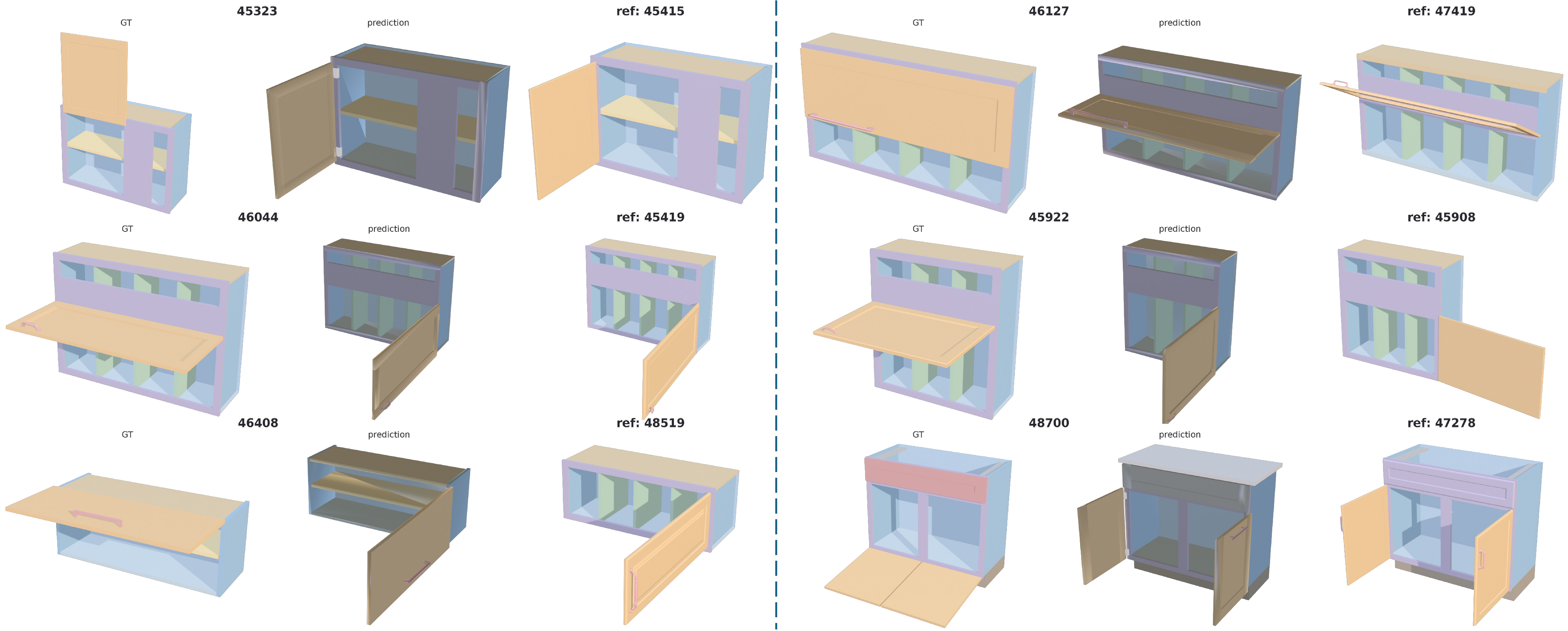}
    \caption{Reasonable motion prediction result showcase. For each grid, left: GT; middle: GraFu prediction; right: reference GT sample used to identify the reasonable motion. Model ID is noted above.}
    \label{fig:mp_reasonable}
\end{figure}

\begin{figure}
    \centering
    \includegraphics[width=\linewidth]{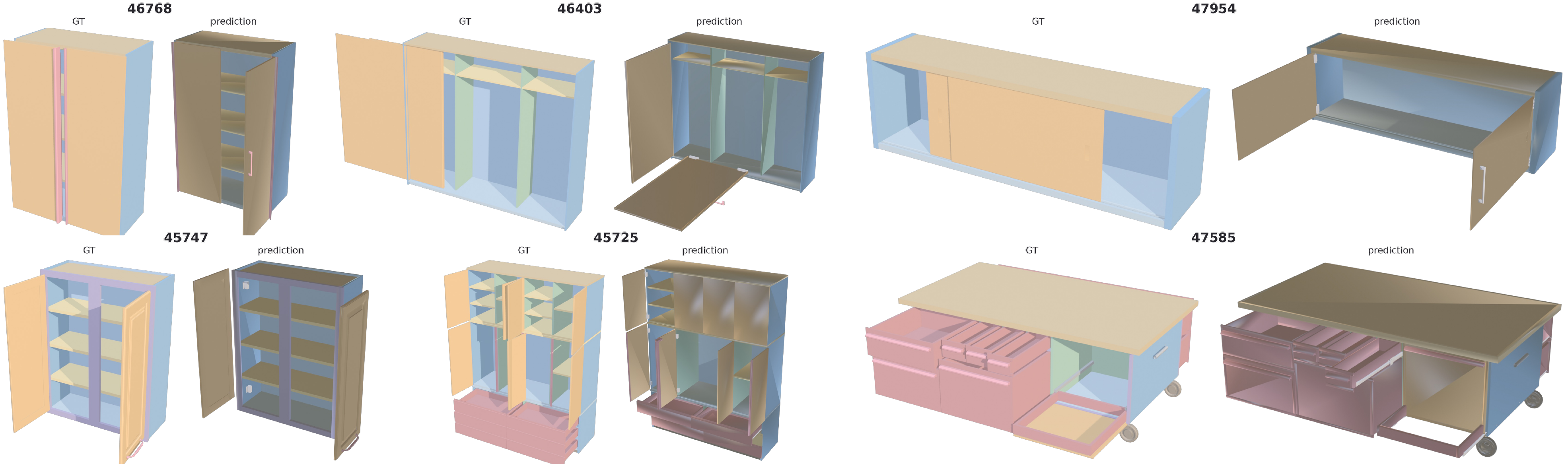}
    \caption{Motion prediction failure cases. Our mechanical joint template gallery does not yet support sliding door motion (Row 1). Other cases include incorrectly formed functional edges between the dynamic node and the mounting static node, leading to incorrect geometric instantiation (Row 2 Column 1); and the high node-count models where some dynamic nodes are left static.}
    \label{fig:mp_failure}
    \vspace{-2 em}
\end{figure}
Of the 345 PN-M test models, 266 (77.1\%) are assigned motions that align with the ground-truth annotation, and a further 62 (18.0\%) receive reasonable motions that differ from the annotation; 17 (4.9\%) are incorrect. \Cref{fig:mp_correct} shows eight randomly selected GT-aligned examples, and \Cref{fig:mp_reasonable} six randomly selected examples that are plausible but not GT-aligned, along with the reference models that allow the prediction to be marked as reasonable.

\Cref{fig:mp_failure} shows five representative failure cases in three groups: motions outside our current taxonomy (e.g.\ sliding doors), incorrect functional-edge predictions, and models with very large node counts, which strain the inference in general.
\subsection{Top Panel Completion}
Missing top panels are the most frequent functional deficiency in PN-M. GraFu detects and repairs them automatically using the prior learnt from the disentangled FurFun-233 dataset, with near-perfect accuracy. \Cref{fig:struct100} shows 100 pairs randomly sampled from the 164 completed models.

\begin{figure*}
    \centering
    \includegraphics[width=0.9\linewidth]{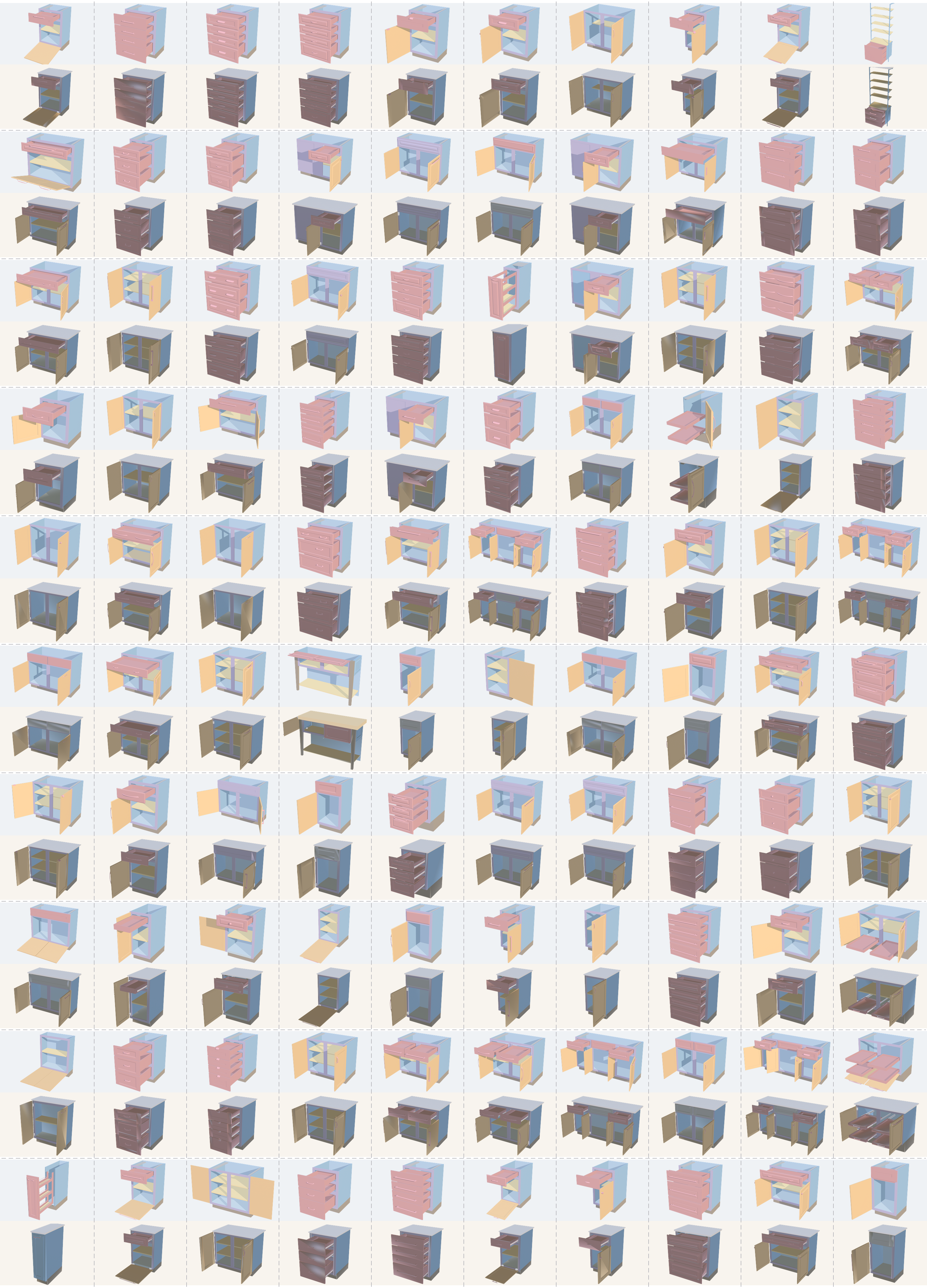}
    \caption{100 randomly sampled PN-M models with missing tops. In each cell, top: input PN-M model; bottom: functionalized model with mechanical parts inserted and top panel added. Please zoom in for details.}
    \label{fig:struct100}
\end{figure*}

\subsection{Handle Insertion}
When handle nodes are already present in the input graph, they act as a prior for motion prediction. The position of a handle indicates how the corresponding movable part actuates. Row~7 of Figure~5 in the main text illustrates this behaviour, where GraFu predicts the door to open away from the handle. This is the more natural motion in practice even though it contradicts the GT annotation. When no handle is present, GraFu predicts the door motion without this cue, and we insert a handle consistent with the predicted axis.

\begin{figure*}
    \centering
    \includegraphics[width=\linewidth]{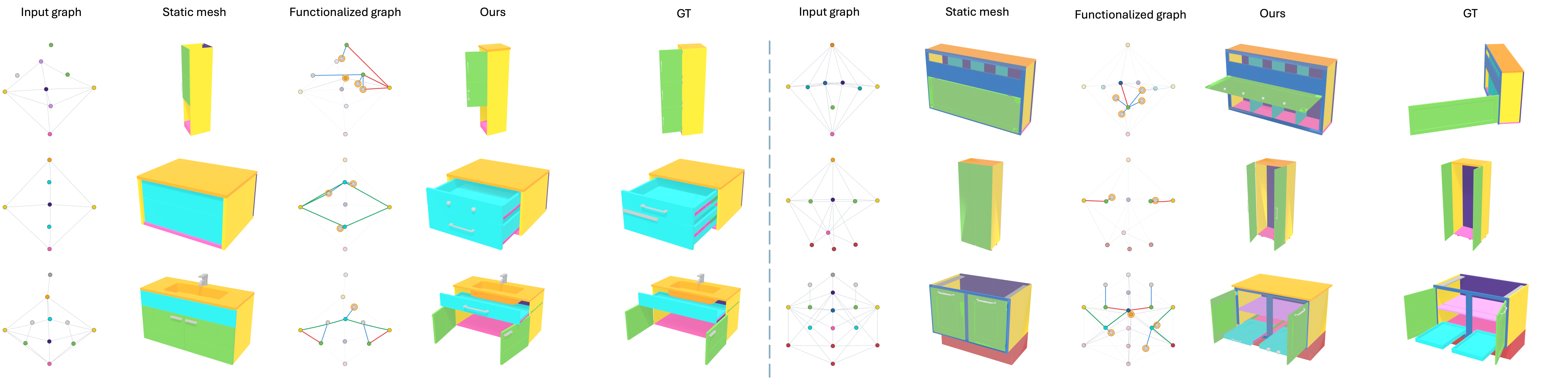}
    \caption{Gallery of handle insertion results. All new handle nodes are predicted by GraFu and installed automatically. Please zoom in for details.}
    \label{fig:handle insertion}
\end{figure*}

\Cref{fig:handle insertion} presents a gallery of newly predicted handle nodes together with their automatic installation. In the Blender add-on, users can further select the handle style, choose an additive or subtractive form, add or remove handles, and adjust handle positions.

We next evaluate handle completion quantitatively on the PN-M test set. We regard a movable part as interactable once it is coupled with at least one handle. On input, 310 (89.9\%) models are fully equipped with additive handle nodes; after GraFu prediction this rises to 329 (95.4\%). The remaining 16 (4.6\%) cases have two causes. First, the model currently issues $K=4$ free-slot queries, which caps the number of handle nodes that can be instantiated on models with many movable parts. Second, cross-distribution generalization: five cases contain recessed drawer units behind a door, a structural and geometric configuration not represented in the training data. FurFun also contains a large proportion of subtractive handles, such as recessed sockets and grooves, so no strong prior that every movable part must carry an additive handle is enforced during training. 

Ideally, the model would probe the geometry of a movable part to determine whether an interactable feature, additive or subtractive, is already present, and insert a new handle only when it is not. The current design does not operate at this geometric granularity, and we regard this as a promising direction for future work. Representative failure cases are shown in \Cref{fig:handle_failure}.

\begin{figure}
    \centering
    \includegraphics[width=0.8\linewidth]{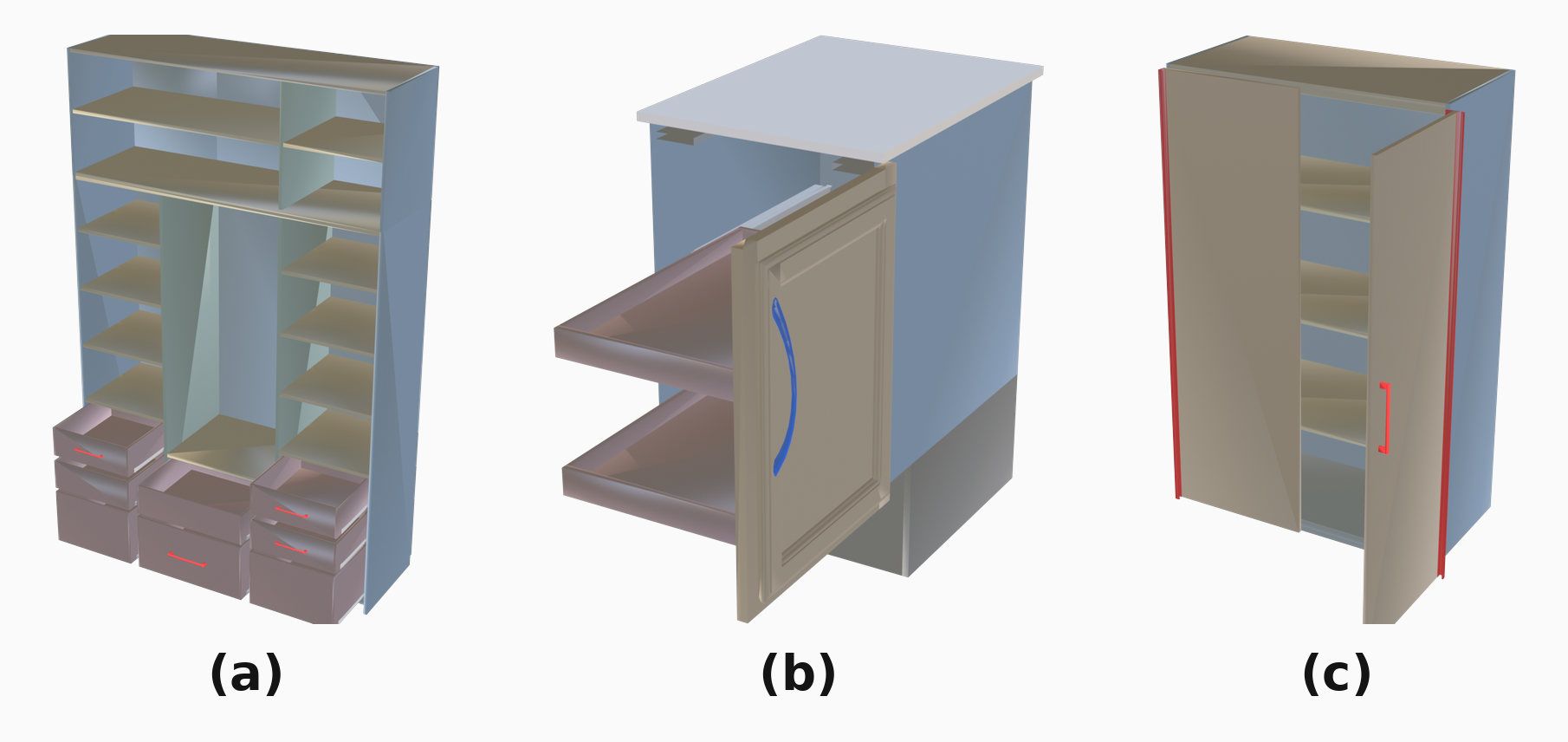}
    \caption{Representative failure cases in handle prediction; existing handles are rendered in blue and newly added handles are rendered in red. (a) The number of required new nodes exceeds the free-slot budget $K=4$. Only 4 new handles are added and the rest are handle-less (b) Unusual, out-of-distribution structures such as recessed drawer units behind a door. The two recessed drawers miss the new handle nodes. (c) Incorrect motion prediction. The static door did not have new handle inserted due to lack of prior motion.}
    \label{fig:handle_failure}
  
\end{figure}

\subsection{Interior Generation}
\begin{figure*}
    \centering
    \includegraphics[width=0.9\linewidth]{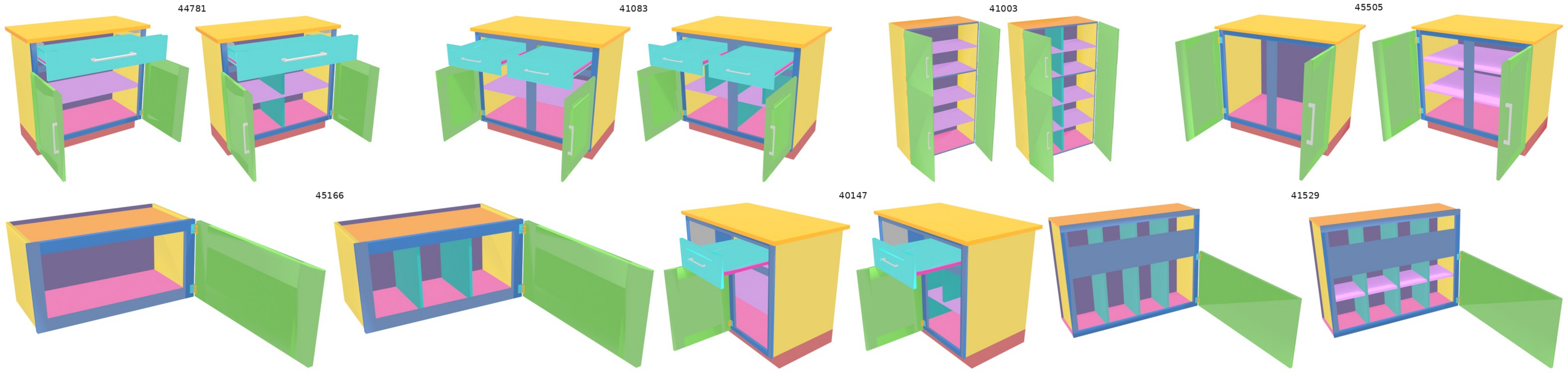}
    \caption{Gallery of interior completion results. In each pair, the original model is shown on the left and the automatic interior completion on the right.}
    \label{fig:interior completion}
\end{figure*}

We expose interior generation as a user-in-the-loop feature, driven entirely by our interior completion algorithm. Figure~6(i) of the main text shows user-designed results obtained through the Blender add-on; below we evaluate the fully automatic setting on PN-M.

The algorithm first detects the compartments available in a model and then inserts divider or shelf geometries into them. We run it on all PN-M test models, inserting one to three dividers or shelves at random into each detected compartment; \Cref{fig:interior completion} shows the resulting completions. Of the 345 models, 77 contain no compartment suitable for insertion and are correctly identified as inapplicable. Among the remaining 268, the algorithm produces reasonable completions for 264 (98.5\%). The four exceptions have irregular compartment shapes that require non-rectilinear geometry (\Cref{fig:interior failure}), which the current geometry library does not yet provide.

\begin{figure}
    \centering
    \includegraphics[width=0.8\linewidth]{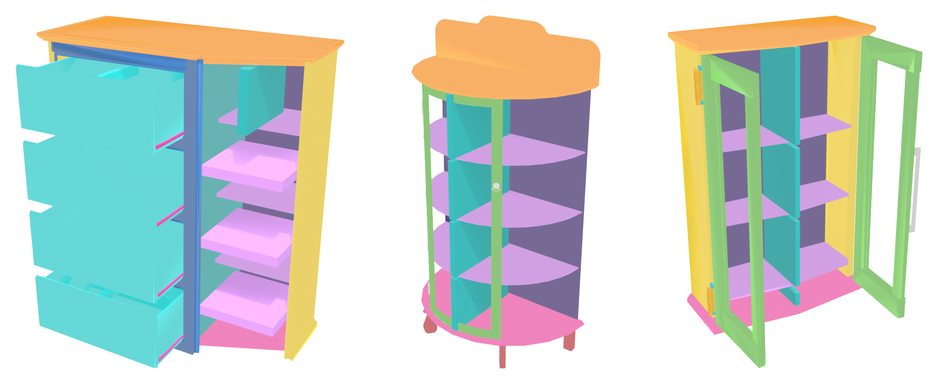}
    \caption{Our interior completion algorithm does not yet support irregularly shaped shelves and dividers, such as circular or slanted ones.}
    \label{fig:interior failure}
\end{figure}

\subsection{Out-of-distribution Graphs}
We further stress-test the pipeline on challenging input graphs, ones that deviate substantially from the training distribution. For each graph, a feature vector is constructed from the total node count, the per-category node counts, the average node degree and the overall geometric bounding box, and measure the disparity between two graphs as the Euclidean distance between their feature vectors.

For each evaluation set, PN-M and HSSD, we report the five test models with the largest z-scored distance from the corresponding training distribution, shown in \Cref{fig:hssd test,fig:pnm test} together with their nearest training-set neighbour on the right. These cases are the most challenging in their respective test sets, and the residual errors are correspondingly concentrated here, namely dynamic nodes that are left static and occasional incorrect motion predictions. Overall, even at extreme node counts and far from the training distribution, the pipeline generalizes well, recovering most functional relations and instantiating the corresponding mechanical joints correctly.

\begin{figure}
    \centering
    \includegraphics[width=\linewidth]{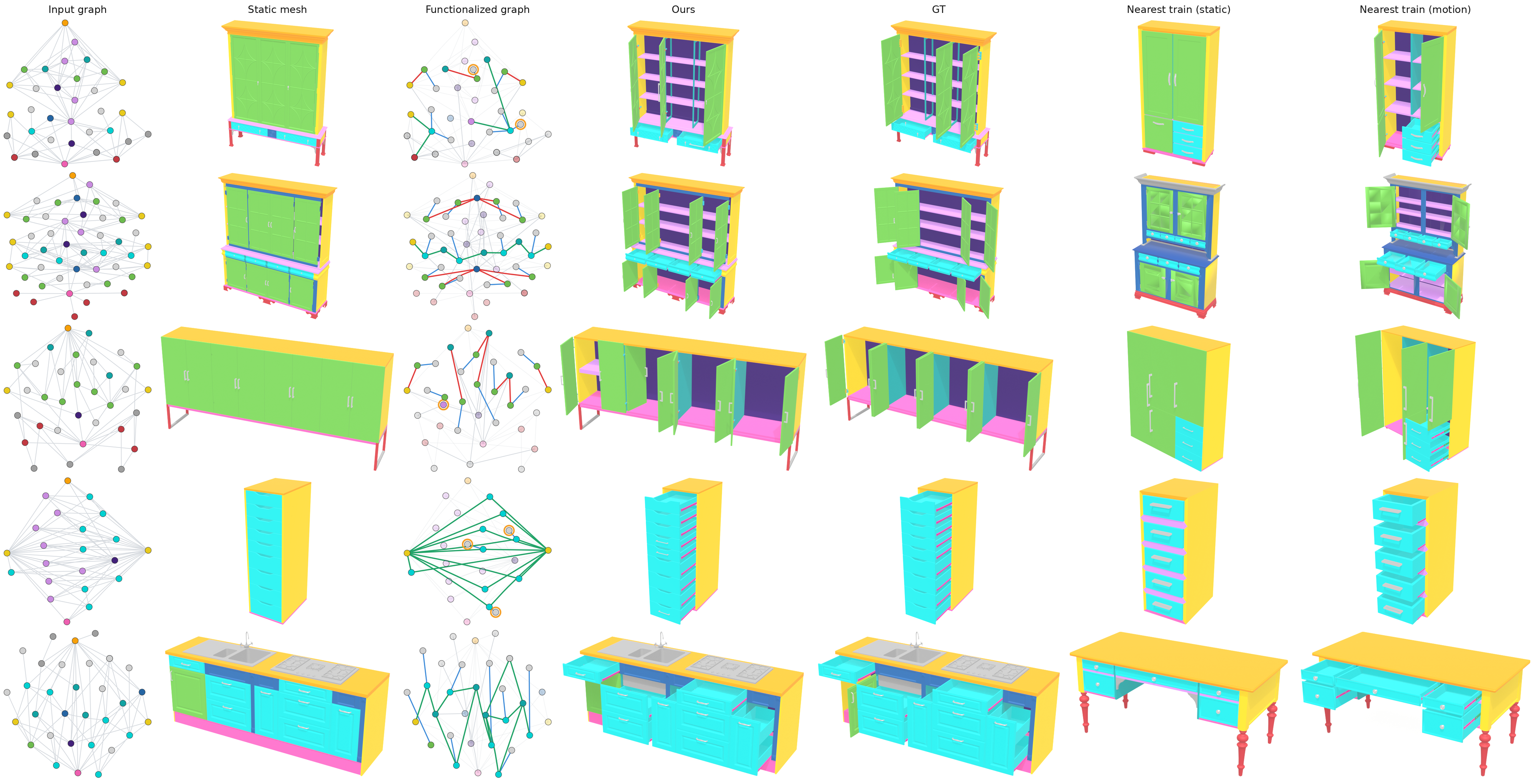}
    \caption{Top five highest-disparity input graphs from the HSSD test set.}
    \label{fig:hssd test}
\end{figure}

\begin{figure}
    \centering
    \includegraphics[width=\linewidth]{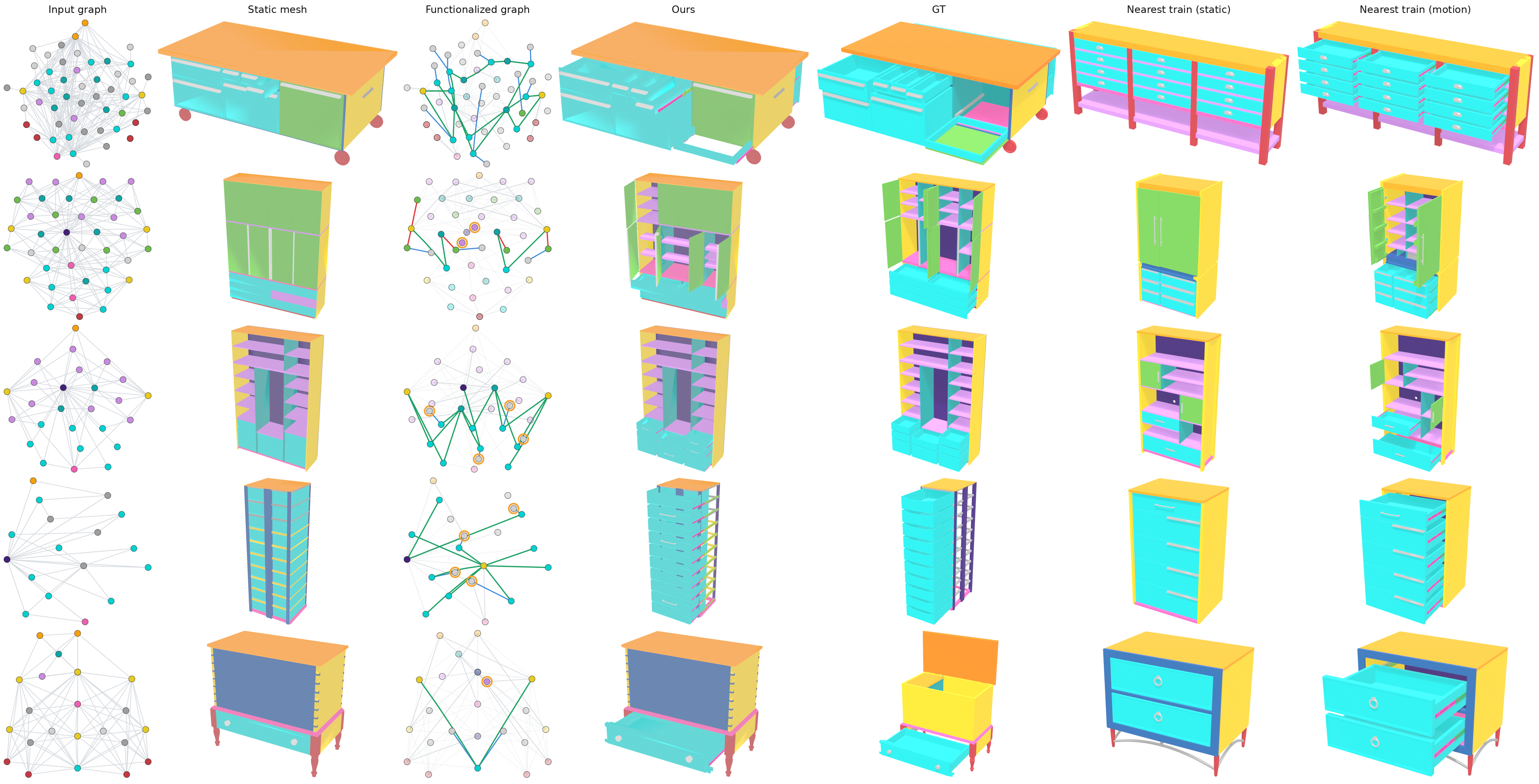}
    \caption{Top five highest-disparity input graphs from the PN-M test set.}
    \label{fig:pnm test}
\end{figure}
\end{revblock}

\section{Blender Add-on}
We provide a Blender add-on that wraps the geometric realization stage into an interactive tool: automatic functionalization first, per-joint refinement after. From a single prediction the user obtains a fully functionalized, animated model in one pass, then inspects, switches, or refines every installed connector and synthesized node directly in the 3D viewport.

\paragraph{Pipeline integration.} The add-on consumes the artifacts produced by inference: the predicted functional graph and the per-part meshes. Loading a model imports the parts coloured by predicted category and automatically installs the predicted hinge, rail, top panel and handles. The interior can be installed on request. The output is an animated \texttt{.blend} file whose doors and drawers
animate through the installed mechanisms over a closed--open--closed cycle, and with structure completed.

\paragraph{Hinges.} Eight templates span the three mechanical classes of C-, P- and F-hinges. Per joint, the user keeps the default collision-tested class selection or forces a class, switches the template variant, and adjusts scale and the number of hinges per door. 

\paragraph{Rails.} Each drawer offers a centre-mount or corner-mount rail pair, optionally backed by support slabs or dividers that bridge the rails to the cabinet walls.

\paragraph{Handles.} Handles are managed per parent door or drawer, including parts for which even no new handle nodes was predicted. The user chooses among 18 templates (bar, knob, and cup styles) or a recessed carved form, and sets the count and placement offsets; multiple handles distribute at equal partitions and shrink automatically to fit. Placement takes the predicted motion as a prior, mounting door handles opposite the hinge axis and centring drawer handles on the front face, so no manual snapping is required. Handles are automatically installed when a new node is predicted, and users can reject the install, or initiate new installations on movable parts that have no new handle nodes predicted.

\paragraph{Tops.} A coverage test reports whether a top panel is missing; a single click then synthesizes one above the silhouette of the body parts, matching the cabinet footprint.

\paragraph{Interior.} Interior structure completion detects compartments that appear as unoccupied boxes, and the user furnishes selected ones with a chosen orientation, panel count, thickness, and front inset. Different compartments can host different configurations within the same session.

\section{Comparison with LLM-based method}
\begin{figure}
    \centering
    \includegraphics[width=\linewidth]{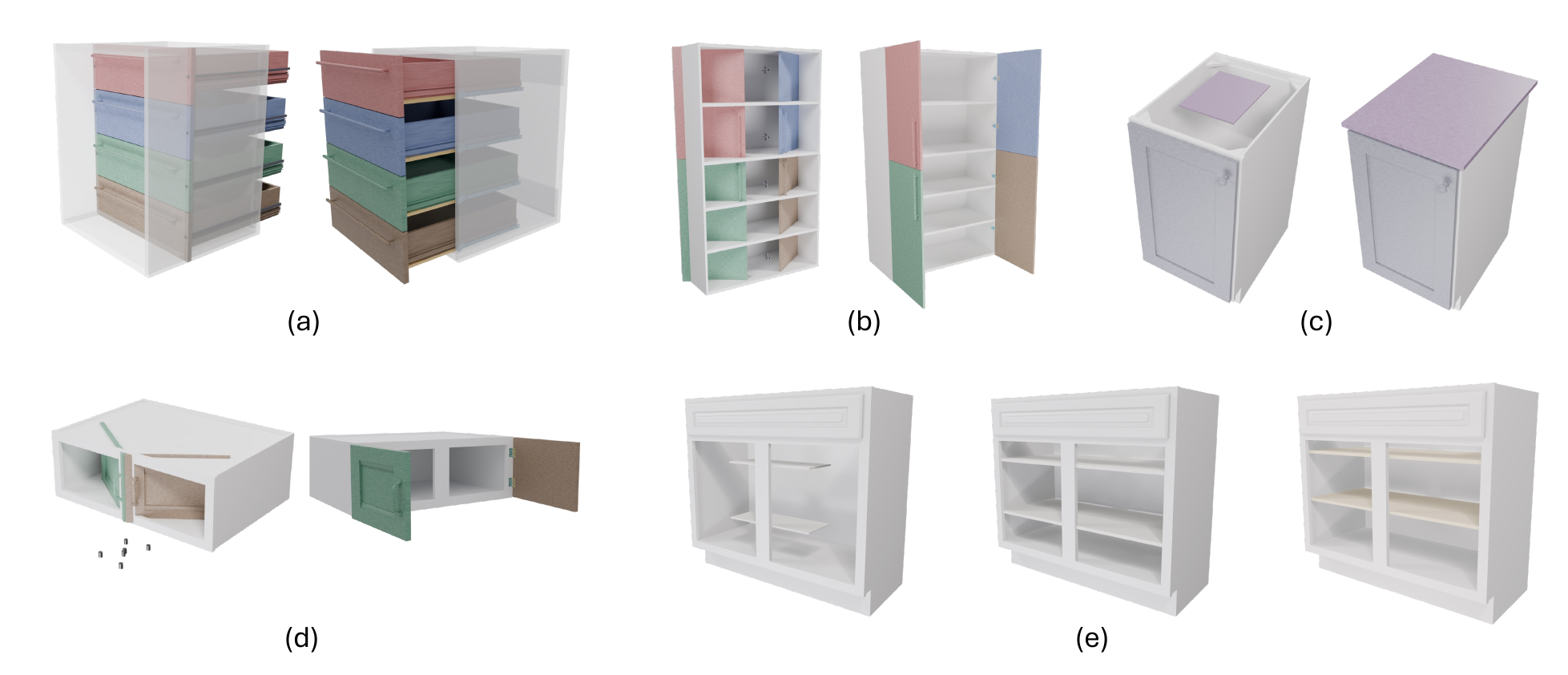}
    \caption{Comparison against BlenderMCP. Left: BlenderMCP, right: Ours. (a) Adding rails to the drawer. (b)(d) Adding hinges to the cabinet. (c) Adding countertops. (e) Adding interior shelves. For each case, multiple prompts are provided from coarse "functionalize it" to detailed step-by-step instructions. (e) shows two BlenderMCP attempts, where the middle case undergoes iterative prompt tuning.}
    \label{fig:mcp_comparison}
\end{figure}

We compare our geometric realization algorithms against BlenderMCP, a tool that connects a vision-language model (VLM) to Blender.
We use Claude 4.5 Sonnet as the VLM, and prompt it to functionalize the object under a range of instructions with different amounts of detail.
We show the results in Figure~\ref{fig:mcp_comparison}.
When prompted only with ``functionalize the object", the VLM fails to apply any meaningful edits. 
We thus exclude these results.
When prompted with the same edit types as ours, the VLM still fails to make accurate edits, leading to inaccurately positioned parts, as well as floating geometry (a-d, left image of e).
Only after additional detailed instructions can the VLM perform satisfactory edits (middle image of e).
In contrast, our method performs the edits consistently.

We further tested on Google Gemini on graph completion results. We provide 20 ground truth functional graphs as the context, and provide detailed instructions in defining nodes, edges and functionality criteria. Upon inference, we provide an unfunctional graph and a model's render, then explicitly query for graph completion. The commercial VLM performed poorly even with sufficient context, with the returned graph containing many isolated nodes.

\bibliographystyle{ACM-Reference-Format}
\bibliography{bib}


\end{document}